\documentclass[letterpaper,10pt,conference]{ieeeconf}

\IEEEoverridecommandlockouts
\usepackage{times}
\usepackage{cite}
\usepackage{booktabs}
\usepackage{multirow}
\usepackage{tabularx}
\usepackage{array}
\usepackage{dcolumn}
\usepackage{graphicx}
\usepackage{amsmath}
\usepackage{amssymb}
\usepackage{mathtools}
\usepackage{makecell}
\usepackage{float}
\usepackage[bookmarks=true,hidelinks]{hyperref}

\newcolumntype{Y}{>{\centering\arraybackslash}X}
\newcolumntype{N}[1]{D{.}{.}{#1}}
\makeatletter
\newcolumntype{B}[1]{>{\boldmath\DC@{.}{.}{#1}}c<{\DC@end}}
\makeatother

\newcommand{\newparagraph}[1]{\noindent{\bfseries #1.}}

\hypersetup{
  pdftitle={Recovering Low-Bit Failures in Vision-Language-Action Policies via Quantization-Conditioned Reinforcement Learning},
  pdfsubject={arXiv manuscript with supplementary appendices}
}

\title{\LARGE \bfseries
Recovering Low-Bit Failures in Vision-Language-Action Policies via Quantization-Conditioned Reinforcement Learning
}

\author{
  \textbf{
  Seongmin Park\textsuperscript{1},
  Jaewon Lee\textsuperscript{1},
  Yujun Heo\textsuperscript{1},
  Yoonsik Kim\textsuperscript{2}
  }
  \\
  \textbf{
  Seheon Choi\textsuperscript{2},
  Hyoung-rock Kim\textsuperscript{2},
  Jungwook Choi\textsuperscript{1,*}\thanks{\textsuperscript{*}Corresponding author: Jungwook Choi (choij@hanyang.ac.kr).}
  }
  \\
  \textsuperscript{1}Hanyang University, Republic of Korea \\
  \textsuperscript{2}LG Electronics, Republic of Korea \\
  \texttt{\{skstjdals, crm06169, creator, choij\}@hanyang.ac.kr} \\
  \texttt{\{yoonsik.kim, ruben.choi, hyoungrock.kim\}@lge.com}
}
\hypersetup{pdfauthor={Seongmin Park, Jaewon Lee, Yujun Heo, Yoonsik Kim, Seheon Choi, Hyoung-rock Kim, Jungwook Choi}}

\makeatletter
\newcommand{\setfigurecaption}{\def\@captype{figure}}
\newcommand{\settablecaption}{\def\@captype{table}}
\makeatother

\makeatletter
\newcommand{\appendixfloatlayout}{%
  \setlength{\@fptop}{0pt}%
  \setlength{\@dblfptop}{0pt}%
  \setlength{\@fpsep}{12pt}%
  \setlength{\@dblfpsep}{12pt}%
}
\makeatother

\begin{document}
\bstctlcite{IEEEcontrol}
\maketitle
\thispagestyle{plain}
\pagestyle{plain}
\begin{abstract}
Quantization lowers deployment costs of vision--language--action (VLA) policies but can compromise task completion. Across five VLA models and 4,050 real-world trials, we systematically reveal that quantization-aware training (QAT) generally improves on post-training quantization (PTQ) but leaves substantial gaps to full precision (FP) on long-horizon tasks and under visual and semantic shifts, while its imitation-based objective does not explicitly use rollout success or partial completion. To exploit these outcome signals, we introduce Quantization-Conditioned Reinforcement Learning (QCRL), an offline RL method coupling progress-based advantage estimation with quantization-aware source conditioning. FP rollouts supply reference experience beyond action-level distillation, progress-scaled terminal penalty distinguishes partial completion, and quantization-aware conditioning selects advantageous reference behavior within the quantized policy. We evaluate weight-only INT4 W4A16 on $\pi_{0.5}$ and GR00T N1.5 and joint weight--activation NVFP4 W4A4 on $\pi_{0.5}$. Using only in-distribution data, QCRL reaches 99.3\% of FP's mean score across three visual-shift conditions for $\pi_{0.5}$ with INT4 W4A16, alongside 42.9\% model-memory savings. With NVFP4 W4A4, it reaches 100.7\% of the same FP baseline, alongside a $1.73\times$ speedup using NVFP4/FP8 deployment. Hardware gains are relative to FP16 on Jetson Thor.

\end{abstract}

\section{Introduction}
\label{sec:introduction}

Vision--language--action (VLA) policies integrate visual observations, language instructions, and action generation, but their scale makes deployment on resource-constrained robots costly~\cite{brohan2023rt2,kim2024openvla,black2024pi0,pi05}. Quantization offers two complementary deployment benefits: lower memory use and faster inference. Weight-only quantization reduces model memory while retaining high-precision activations~\cite{frantar2022gptq}, as in Core ML's 4-bit weight compression with FP16 activations on the Apple Neural Engine~\cite{appleweightcompression}. Joint weight--activation quantization also accelerates inference through low-precision matrix multiplication~\cite{nagel2021whitepaper}; NVFP4 W4A4 enables 4-bit weight and activation computation on Blackwell devices such as Jetson Thor~\cite{nvidiajetsonthor}.

These efficiency gains motivate quantized deployment, but preserving successful robot control remains a concern: action deviations can alter contact, subsequent observations, and task completion. To identify where quantization compromises execution and where QAT recovery falls short, we conduct a comprehensive comparative study of full-precision (FP), post-training quantization (PTQ), and quantization-aware training (QAT) policies across 4,050 real-world trials and five VLA models~\cite{park2025sqil}. Despite strong recovery on short tasks, QAT leaves severe long-horizon completion failures and substantial visual and semantic out-of-distribution (OOD) gaps due to QAT's shortsighted step-wise guidance. These findings motivate learning from long-term rollout success and partial completion, signals not explicitly used by the evaluated imitation-based QAT objective.

\begin{figure*}[t]
\centering
\includegraphics[width=\textwidth]{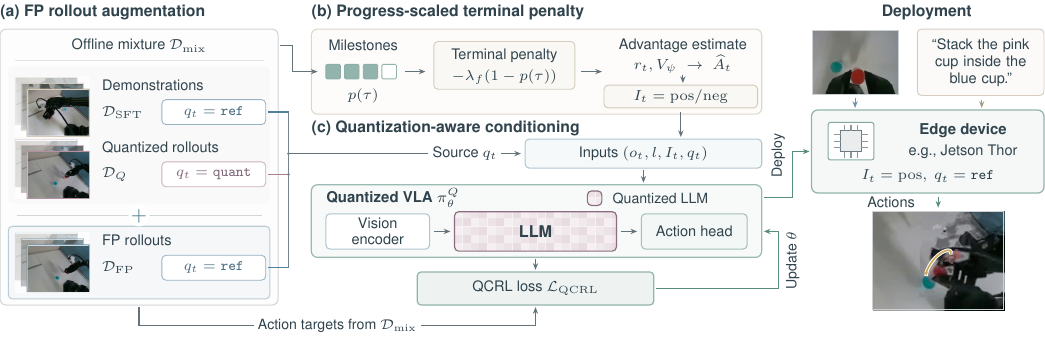}
\vspace{-0.6cm}
\caption{QCRL components: (a) FP rollout augmentation; (b) progress-scaled terminal penalty and advantage labels; (c) quantization-aware conditioning and policy learning. Dataset actions provide targets for $\mathcal{L}_{QCRL}$ (Eq. (8)); deployment selects the positive reference condition.}
\label{fig:qcrl_overview}
\end{figure*}

To embody these findings for low-bit recovery, we propose Quantization-Conditioned Reinforcement Learning (QCRL), a new offline RL method coupling progress-based advantage estimation with source-conditioned behavior selection (Fig.~\ref{fig:qcrl_overview}). \emph{FP rollout augmentation} collects reference trajectories beyond action-level distillation, without new expert demonstrations. A \emph{progress-scaled terminal penalty} incorporates partial completion into advantage estimates, differentiating failed rollouts. \emph{Quantization-aware conditioning} distinguishes reference and quantized behavior, making advantageous reference behavior separately selectable. 

We evaluate INT4 W4A16 on $\pi_{0.5}$ and GR00T N1.5, and NVFP4 W4A4 on $\pi_{0.5}$. Across appearance, illumination, and sensor-degradation shifts, in-distribution (ID)-only QCRL adaptation raises $\pi_{0.5}$'s mean task score from QAT's 77.5\% to 99.3\% of FP under W4A16, alongside 42.9\% model-memory savings. Under NVFP4 W4A4, the corresponding score rises from 87.0\% to 100.7\% of FP, while NVFP4/FP8 deployment achieves a $1.73\times$ speedup. Hardware gains are measured against FP16 on Jetson Thor. Ablations and an FP adaptation control assess component contributions and recovery beyond general adaptation.

\section{Background and Related Work}
\label{sec:related_work}

\subsection{Quantized VLA Policies}
VLA policies combine vision and language for robotic control. OpenVLA, $\pi_0$, $\pi_{0.5}$, $\pi_0$-FAST, SmolVLA, and GR00T vary in backbone, action representation, generation head, and adaptation procedure~\cite{kim2024openvla,black2024pi0,pi05,pertsch2025fast,shukor2025smolvla,grootn1}. Their memory and computation demands motivate quantization for on-device deployment~\cite{park2025sqil,zhang2026quantvla}.

Quantization maps a weight or activation $x$ to a low-bit representation $x^Q$, reconstructed as $\widetilde{x}$:
\[
x^Q=\mathcal{R}_{\mathcal{F}}(x/s+z),
\qquad
\widetilde{x}=s(x^Q-z),
\]
where $s>0$ is the scale, $z$ the zero point, and $\mathcal{R}_{\mathcal{F}}$ rounds to the nearest value in format $\mathcal{F}$, saturating at its limits.

Weight-only quantization reduces parameter storage and memory traffic~\cite{frantar2022gptq}; INT4 W4A16 uses uniform four-bit integer weights and 16-bit floating-point activations. ONNX Runtime defaults to round-to-nearest (RTN) for blockwise four-bit MatMul weight quantization~\cite{onnxruntimequantization}. Joint weight--activation quantization also enables low-precision matrix multiplication on supported hardware. NVFP4 W4A4 uses E2M1 weights and activations, with $z=0$, FP8 block scales, and an FP32 tensor scale~\cite{tensorrtquanttypes,cutlassblackwell}. Blackwell-based Jetson Thor supports FP4 Tensor Core computation for efficient deployment~\cite{nvidiajetsonthor}.

Post-training quantization (PTQ) converts trained parameters without retraining but can degrade accuracy. VLA-specific methods include QuantVLA's scale calibration and QVLA's action-sensitive bit allocation~\cite{zhang2026quantvla,xu2026qvla}. Quantization-aware training (QAT) can recover accuracy more effectively than PTQ by adapting parameters under quantized forward passes~\cite{jacob2018quantization,nagel2021whitepaper}. Saliency-aware quantized imitation learning (SQIL) adds saliency-weighted FP distillation, improving robustness over PTQ in evaluated robotic settings~\cite{park2025sqil}. However, systematic evaluation of PTQ and QAT across diverse VLA models, real-world task difficulties, and distribution shifts remains limited. In Sec.~\ref{sec:comparative_study}, we evaluate SQIL and calibration-free RTN across five VLA models under a shared real-world protocol to characterize QAT's recovery and its limits.

\subsection{Offline Policy Improvement}
\label{sec:recap_preliminaries}
Offline policy improvement reuses demonstrations and rollout outcomes without new robot interactions during optimization, while accounting for data--policy mismatch~\cite{levine2020offline}. Advantage-weighted regression (AWR) emphasizes actions with higher estimated advantage~\cite{peng2019awr}, while implicit Q-learning (IQL) avoids evaluating unseen actions and extracts policies through advantage-weighted behavioral cloning~\cite{kostrikov2021iql}. RECAP improves VLAs from demonstrations, autonomous rollouts, and human corrections through advantage conditioning~\cite{amin2025pi06}. We summarize its policy-update mechanism on a fixed dataset, which provides the background for QCRL.

Let $\mathcal D$ contain trajectories $\tau$ with observations $o_t$, instruction $l$, actions $a_t$, and final transition index $T(\tau)$. RECAP assigns $r_t=-1$ at non-terminal steps and a terminal reward based on task success:
\begin{equation}
\label{eq:recap_reward}
r_{T(\tau)}=
\begin{cases}
0, & \text{success},\\
-\lambda_f, & \text{failure},
\end{cases}
\end{equation}
where $\lambda_f>0$ is the terminal failure-penalty weight.

A value model fitted to return-to-go targets $\widehat R_t=\sum_{j=t}^{T(\tau)}\gamma^{j-t}r_j$ provides the estimate $V_\psi(o_t,l)$, where $\gamma$ is the discount factor (RECAP uses $\gamma=1$). For non-terminal segments of $N$ steps, the estimated advantage is
\begin{equation}
\label{eq:qcrl_advantage}
\widehat A_t
=\sum_{j=0}^{N-1}\gamma^j r_{t+j}
+\gamma^N V_\psi(o_{t+N},l)-V_\psi(o_t,l).
\end{equation}
Thresholding this estimate gives $I_t=\mathrm{pos}$ if
$\widehat A_t\geq\epsilon_\ell$ and $I_t=\mathrm{neg}$
otherwise, where $\epsilon_\ell$ is a task-specific threshold.

For policy $\pi_\theta$ with trainable parameters $\theta$, the rescaled advantage-conditioned objective is
\begin{equation}
\label{eq:advantage_conditioned_learning}
\begin{aligned}
\mathcal L_{\mathrm{AC}}(\theta)
=\mathbb E_{\mathcal D}\bigl[
&\ell_{\pi_\theta}(a_t\mid o_t,l,I_t)\\
&+\beta_u\ell_{\pi_\theta}(a_t\mid o_t,l)\bigr],
\end{aligned}
\end{equation}
where $\ell_{\pi_\theta}$ is the action-modeling loss and $\beta_u$ weights its auxiliary term without advantage conditioning. Training uses both label groups; deployment selects $I_t=\mathrm{pos}$.

The residual performance gaps examined in our comparative study motivate offline policy improvement for quantized VLAs, but this approach raises a further question: how should offline improvement use experience collected at different precisions to recover a policy that remains quantized at deployment? QCRL addresses this question by adapting advantage-conditioned learning to incorporate behavior from different sources and partial task progress (Sec.~\ref{sec:method}).

\begin{figure*}[t]
\centering
\includegraphics[width=\textwidth]{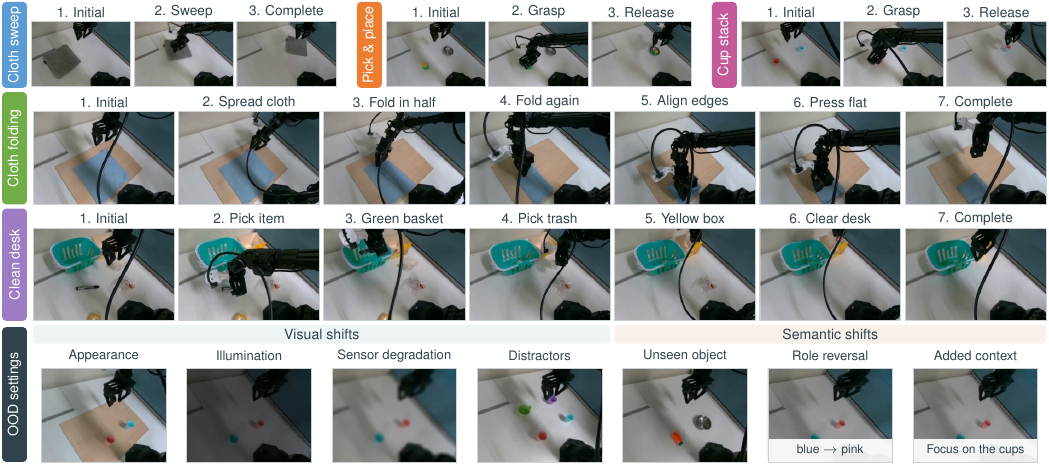}
\vspace{-0.4cm}
\caption{Real-world task sequences (left to right): short tasks, folding, and clean desk in rows 1--3. Bottom row: visual and semantic OOD conditions.}
\label{fig:task_overview}
\end{figure*}

\section{Comparative Study of Quantized VLAs}
\label{sec:comparative_study}

Reliable quantized VLA deployment requires systematically assessing quantization's effects across VLA models and operating conditions. We compare FP, PTQ, and QAT in 4,050 real-world trials across five VLA models to measure performance degradation, QAT recovery, and remaining gaps.

\subsection{Study Design}
\label{sec:comparative_protocol}

We evaluate $\pi_0$, $\pi_{0.5}$, $\pi_0$-FAST, SmolVLA, and GR00T N1.5, covering different backbones and action-generation designs~\cite{black2024pi0,pi05,pertsch2025fast,shukor2025smolvla,grootn1}. For each model, its BF16 task-trained checkpoint provides the FP reference and initialization for RTN PTQ and SQIL QAT~\cite{park2025sqil}. Both quantize language-model weights to INT4 with group size 128, retaining 16-bit activations and the original precision of vision and action modules. Table~\ref{tab:baseline_training} gives training schedules.

\begin{table}[!htbp]
\centering
\caption{FP/QAT training: steps / learning rate ($10^{-5}$). openpi (JAX): $\pi$ models~\cite{openpirepo}; LeRobot: SmolVLA/GR00T~\cite{cadene2024lerobot}.}
\label{tab:baseline_training}
\footnotesize
\setlength{\tabcolsep}{1.3pt}
\renewcommand{\arraystretch}{1.1}
\begin{tabular*}{\linewidth}{@{\extracolsep{\fill}}lcccc@{}}
\toprule
Setting & $\pi_0$, $\pi_{0.5}$ & $\pi_0$-FAST & SmolVLA & GR00T N1.5 \\
\midrule
Chunk/batch & 50 / 2 & 32 / 2 & 50 / 64 & 16 / 32 \\
FP  & 100k / 2.5 & 140k / 5 & 3.2k / 20 & 16k / 10 \\
QAT & 50k / 2 & 70k / 2 & 1.6k / 20 & 8k / 10 \\
\bottomrule
\end{tabular*}
\end{table}

\subsection{Tasks and metrics}
\label{sec:task_protocol}

We evaluate sweeping, pick-and-place, stacking, folding, and desk cleaning on a Trossen ViperX follower arm with base/wrist cameras. Visual and semantic shifts modify observations, objects, or instructions (Fig.~\ref{fig:task_overview}). Each model--method pair uses 30 trials per short ID task and 10 per long task or OOD task--condition combination.

Full success requires completing the entire task. To distinguish partial execution from failure without progress, we also record completed milestones: picking/placing for pick-and-place and stacking, each fold for folding ($K=2$), and each correctly placed object for clean desk ($K=4$). The progress score is
\begin{equation}
\label{eq:progress_score}
S_{\mathrm{progress}}=\frac{1}{K}\sum_{k=1}^{K}\frac{n_k}{M},
\end{equation}
where $K$ is the milestone count, $M$ the trial count, and $n_k$ the number reaching milestone $k$. This measures mean trajectory completion; for semantic shifts, primitive progress tracks task-specific action stages. All scores are percentages.

\subsection{Limits of QAT Recovery}
\label{sec:structured_sensitivity}
\label{sec:distribution_shift_results}

\begin{figure*}[t]
\centering
\includegraphics[width=\textwidth]{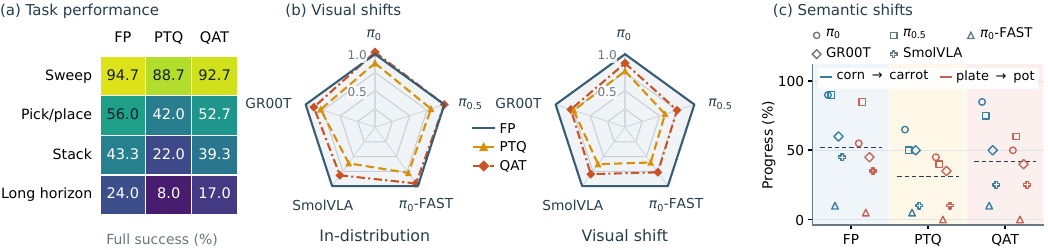}
\vspace{-0.7cm}
\caption{Quantization effects across five VLA models. (a) Mean full success rate; long horizon averages folding and clean desk. (b) ID averages three tasks; visual averages 11 task--condition scores. Each mean is normalized to the model's corresponding FP mean ($\mathrm{FP}=1$), retaining ratios above one. (c) Primitive progress scores for object and recipient replacement; markers identify models, and dashed lines average all five models across both shifts.}
\vspace{-0.2cm}
\label{fig:difficulty_visualgen}
\label{fig:semantic_results}
\end{figure*}

Table~\ref{tab:comparative_models} reports PTQ and QAT task performance alongside each model’s FP reference, first on ID tasks and then under distribution shifts, and Fig.~\ref{fig:difficulty_visualgen} summarizes the key findings.

\begin{table*}[!t]
\centering
\caption{Real-world task performance and OOD generalization across five VLA models (\%). PTQ and QAT use INT4 W4A16.}
\vspace{-0.2cm}
\label{tab:comparative_models}
\footnotesize
\setlength{\tabcolsep}{3pt}
\renewcommand{\arraystretch}{1.08}
\begin{tabular*}{\textwidth}{@{\extracolsep{\fill}}ll*{15}{N{3.1}}@{}}
\toprule
Condition & Metric & \multicolumn{3}{c}{$\pi_0$} & \multicolumn{3}{c}{$\pi_{0.5}$} & \multicolumn{3}{c}{$\pi_0$-FAST} & \multicolumn{3}{c}{SmolVLA} & \multicolumn{3}{c}{GR00T N1.5} \\
\cmidrule(lr){3-5}\cmidrule(lr){6-8}\cmidrule(lr){9-11}\cmidrule(lr){12-14}\cmidrule(lr){15-17}
 & & \multicolumn{1}{c}{FP} & \multicolumn{1}{c}{PTQ} & \multicolumn{1}{c}{QAT} & \multicolumn{1}{c}{FP} & \multicolumn{1}{c}{PTQ} & \multicolumn{1}{c}{QAT} & \multicolumn{1}{c}{FP} & \multicolumn{1}{c}{PTQ} & \multicolumn{1}{c}{QAT} & \multicolumn{1}{c}{FP} & \multicolumn{1}{c}{PTQ} & \multicolumn{1}{c}{QAT} & \multicolumn{1}{c}{FP} & \multicolumn{1}{c}{PTQ} & \multicolumn{1}{c}{QAT} \\
\midrule
Sweep & Full & 96.7 & 100.0 & 100.0 & 100.0 & 100.0 & 100.0 & 83.3 & 80.0 & 83.3 & 93.3 & 70.0 & 83.3 & 100.0 & 93.3 & 96.7 \\
Pick/place & Full & 83.3 & 73.3 & 86.7 & 76.7 & 63.3 & 76.7 & 40.0 & 26.7 & 36.7 & 30.0 & 16.7 & 23.3 & 50.0 & 30.0 & 40.0 \\
Stack & Full & 63.3 & 40.0 & 63.3 & 83.3 & 46.7 & 83.3 & 26.7 & 10.0 & 23.3 & 23.3 & 3.3 & 13.3 & 20.0 & 10.0 & 13.3 \\
\midrule
Folding & Full & 70.0 & 10.0 & 60.0 & 40.0 & 0.0 & 0.0 & 0.0 & 0.0 & 0.0 & 0.0 & 0.0 & 0.0 & 40.0 & 30.0 & 30.0 \\
Folding & Progress & 80.0 & 25.0 & 80.0 & 60.0 & 15.0 & 40.0 & 15.0 & 0.0 & 5.0 & 10.0 & 0.0 & 0.0 & 55.0 & 40.0 & 45.0 \\
Clean desk & Full & 40.0 & 20.0 & 50.0 & 50.0 & 20.0 & 30.0 & 0.0 & 0.0 & 0.0 & 0.0 & 0.0 & 0.0 & 0.0 & 0.0 & 0.0 \\
Clean desk & Progress & 67.5 & 37.5 & 62.5 & 70.0 & 40.0 & 52.5 & 12.5 & 2.5 & 5.0 & 5.0 & 0.0 & 0.0 & 25.0 & 12.5 & 15.0 \\
\midrule
Appearance & Score & 85.0 & 68.3 & 78.3 & 86.7 & 73.3 & 83.3 & 40.0 & 26.7 & 33.3 & 38.3 & 30.0 & 36.7 & 51.7 & 41.7 & 50.0 \\
Illumination & Score & 71.7 & 63.3 & 65.0 & 73.3 & 46.7 & 53.3 & 33.3 & 23.3 & 28.3 & 66.7 & 46.7 & 56.7 & 63.3 & 55.0 & 55.0 \\
Sensor & Score & 55.0 & 33.3 & 41.7 & 70.0 & 13.3 & 41.7 & 26.7 & 10.0 & 15.0 & 43.3 & 16.7 & 25.0 & 45.0 & 21.7 & 20.0 \\
Distractors & Score & 5.0 & 0.0 & 5.0 & 75.0 & 42.5 & 47.5 & 0.0 & 0.0 & 0.0 & 0.0 & 0.0 & 0.0 & 0.0 & 0.0 & 0.0 \\
\midrule
Unseen object & Progress & 72.5 & 55.0 & 67.5 & 87.5 & 45.0 & 67.5 & 7.5 & 2.5 & 5.0 & 40.0 & 10.0 & 25.0 & 52.5 & 42.5 & 45.0 \\
Role reversal & Progress & 20.0 & 30.0 & 26.7 & 43.3 & 43.3 & 40.0 & 10.0 & 6.7 & 6.7 & 43.3 & 6.7 & 20.0 & 33.3 & 33.3 & 30.0 \\
Added context & Progress & 45.0 & 40.0 & 37.5 & 70.0 & 37.5 & 52.5 & 5.0 & 0.0 & 5.0 & 30.0 & 15.0 & 25.0 & 42.5 & 37.5 & 37.5 \\
\bottomrule
\end{tabular*}
\par\smallskip
\begin{minipage}{\textwidth}
\footnotesize
\textit{Metrics.} Full: full success rate; Progress: milestone/primitive progress score; Sensor: sensor degradation.
Visual Score averages sweeping success and pick-and-place/stacking milestone progress (two tasks for distractors).
Unseen object: mean pick/place completion across object/recipient replacements.
Role reversal: mean move-to-target/pick/place completion, regardless of object role.
Added context: mean score for the specified manipulation across preliminary-action and task-focus prompts.
\end{minipage}
\vspace{-0.2cm}
\end{table*}

\newparagraph{Recovery on short tasks}
Sweeping is relatively robust to quantization, while precise placement is more sensitive. For $\pi_{0.5}$ stacking, PTQ reduces success from 83.3\% to 46.7\%, and QAT restores 83.3\%. This recovery varies across models: $\pi_0$-FAST, SmolVLA, and GR00T remain below their FP baselines on both pick-and-place and stacking after QAT.

\newparagraph{Limits on long-horizon completion}
Longer tasks expose failures even in $\pi_{0.5}$, whose short-task scores recover to FP levels. On folding, QAT increases progress from PTQ's 15.0\% to 40.0\%, yet full success remains 0.0\%, versus 40.0\% for FP. QAT thus recovers some task stages without restoring the ability to complete both folds.

\newparagraph{Limits under visual shifts}
Changing observations exposes additional gaps on short tasks: all five QAT policies retain less of their FP performance under visual shifts than under ID conditions. Under sensor degradation, QAT scores are 41.7\% versus FP's 70.0\% for $\pi_{0.5}$, and 20.0\% versus 45.0\% for GR00T. These residual gaps indicate incomplete QAT recovery, while low FP distractor scores also indicate baseline limitations.

\newparagraph{Limits under object and instruction shifts}
Short-task recovery also fails to transfer fully to changed objects and instructions. Object and recipient replacements leave all five QAT policies below FP, with mean primitive progress of 42.0\% versus 52.0\%. For $\pi_{0.5}$, QAT retains gaps of 20.0 percentage points under replacement and 17.5 under added context despite matching FP on short ID tasks. These metrics assess manipulation progress under semantic shifts.

\subsection{Implications for Offline Policy Improvement}
QAT leaves severe deficits in long-horizon completion and substantial gaps under visual and semantic shifts. SQIL trains on recorded ID observations using action supervision and FP-output distillation, but does not explicitly use whether the quantized policy completes the task or completes only some stages. Offline RL can use these outcomes to estimate action advantages from successful, failed, and partially completed trajectories. This motivates QCRL, which uses rollout feedback to improve task execution while retaining low-bit inference.

\section{Quantization-Conditioned Reinforcement Learning}
\label{sec:method}

To address the residual failures in Section~\ref{sec:comparative_study}, QCRL extends RECAP's advantage-conditioned learning through three changes: FP rollouts augment the reference data, partial completion refines the terminal reward, and a source condition makes reference behavior selectable within a quantized policy (Fig.~\ref{fig:qcrl_overview}). QCRL couples progress-based advantage labeling with source-conditioned policy extraction, enabling a quantized policy to learn from deployment failures while selecting reference behavior at inference.

\subsection{FP Rollout Augmentation}
\label{sec:fp_rollout_augmentation}

Let $\pi_{\mathrm{FP}}^0$ be the task-trained FP policy and $\pi_Q^0$ its initial PTQ counterpart. Quantized rollouts $\mathcal D_Q$ capture $\pi_Q^0$'s deployment states and execution errors, thus failure examples may overwhelm. To augment successful experience, we supplement expert demonstrations $\mathcal D_{\mathrm{SFT}}$ with rollouts $\mathcal D_{\mathrm{FP}}$ from $\pi_{\mathrm{FP}}^0$. Together they form the reference dataset $\mathcal D_{\mathrm{ref}}=\mathcal D_{\mathrm{SFT}}\cup\mathcal D_{\mathrm{FP}}$ and the fixed training mixture
\begin{equation}
\label{eq:qcrl_mixture}
\mathcal{D}_{\mathrm{mix}}
=\mathcal{D}_{\mathrm{SFT}}\cup\mathcal{D}_{\mathrm{FP}}\cup\mathcal{D}_{Q}.
\end{equation}

SQIL uses FP outputs as distillation targets, whereas QCRL collects FP rollouts as offline RL data, using an FP policy with the same backbone, action
representation, and task training. This adds reference trajectories that grant high proximity without new expert
demonstrations. Section~\ref{sec:source_selection_analysis} validates the benefits of high proximity to rollout quality.

\subsection{Progress-Scaled Terminal Penalty}
\label{sec:partial_success_reward}

RECAP assigns the same terminal penalty to every failed rollout (Eq.~\eqref{eq:recap_reward}). Yet low-bit failures can occur after useful execution: a folding policy may complete the first fold and fail on the second. A binary terminal outcome disregards this progress in the failure penalty. For each trajectory in $\mathcal D_{\mathrm{mix}}$, let $p(\tau)\in[0,1]$ be the fraction of required subgoals completed. Different from RECAP, QCRL defines the terminal reward as:
\begin{equation}
\label{eq:qcrl_reward}
r_{T(\tau)}=-\lambda_f(1-p(\tau)).
\end{equation}
For example, completing three of four subgoals incurs $-\lambda_f/4$ rather than $-\lambda_f$; full completion receives zero terminal reward. Thus, failed rollouts can receive higher returns when they complete more subgoals. Progress adjusts the terminal penalty rather than supplying a dense reward at every subgoal.

Using this progress-scaled terminal penalty, we fit $V_\psi$ to these return-to-go targets and compute $N$-step advantages and labels $I_t$ as in Section~\ref{sec:recap_preliminaries}. The value targets now distinguish degrees of task completion even when full successes are scarce, allowing partial execution to inform policy improvement. Per-transition advantages still determine the labels, so a partially successful trajectory is not automatically labeled positive throughout.

\subsection{Quantization-Aware Conditioning}
\label{sec:precision_aware_conditioning}
\label{sec:qcrl_objective}

RECAP's advantage label groups actions by estimated advantage
but does not identify their source: both reference and
quantized rollouts can receive positive labels. Without source conditioning, their actions are learned under the same advantage condition; thus, a naive application of RECAP would assign the same positive labels for both reference and quantized rollouts. For quantization-aware source conditioning, QCRL adds a separate source label within each advantage group:
\begin{equation}
\label{eq:precision_source_assignment}
q_t=
\begin{cases}
\mathtt{ref}, &(o_t,l,a_t)\in\mathcal{D}_{\mathrm{ref}},\\
\mathtt{quant}, &(o_t,l,a_t)\in\mathcal{D}_{Q}.
\end{cases}
\end{equation}
Here, $q_t$ identifies data provenance, including demonstrations in the reference group; it does not change the model's execution precision. The same quantized policy $\pi_\theta^Q$ learns from both sources using
\begin{equation}
\label{eq:qcrl_policy_loss}
\begin{aligned}
\mathcal{L}_{\mathrm{QCRL}}(\theta)
=\mathbb{E}_{\mathcal D_{\mathrm{mix}}}\Bigl[
&\ell_{\pi_\theta^Q}(a_t\mid o_t,l,I_t,q_t)\\
&+\beta_u\ell_{\pi_\theta^Q}(a_t\mid o_t,l)\Bigr].
\end{aligned}
\end{equation}
Here, $\ell_{\pi_\theta^Q}$ is the action-modeling loss and $\beta_u$ weights its auxiliary term without advantage or source conditioning. The two labels serve complementary roles: $I_t$ identifies actions with higher estimated advantage, while $q_t$ specifies which source's behavior to model. Quantized rollouts contribute deployment observations and actions to the shared policy, while the reference condition makes behavior learned from demonstrations and FP rollouts separately selectable. Combining both labels selects high-advantage reference behavior without assuming that every reference action is preferable.

At deployment, we fix both conditions to select the positive reference behavior:
\begin{equation}
\label{eq:qcrl_deploy_policy}
\pi_{\mathrm{deploy}}(a_t\mid o_t,l)
=\pi_\theta^Q(a_t\mid o_t,l,I_t=\mathrm{pos},q_t=\mathtt{ref}).
\end{equation}
This quantized policy deployment does not incur inference overhead; the FP policy, value model, and progress annotations are used only during offline training. Detailed data and optimization settings appear in Section~\ref{sec:experimental_setup}.
\begin{table}[!t]
\centering
\caption{QCRL diagnostics. (a): $\pi_{0.5}$ W4A16, mean $\pm$ episode SD. PTQ uses no $q_t$. (c): mean $\pm$ observation SD.}
\vspace{-0.2cm}
\label{tab:source_distance_proxy}
\label{tab:source_diagnostics}
\footnotesize
\setlength{\tabcolsep}{2pt}
\renewcommand{\arraystretch}{1.06}
\begin{tabular*}{\columnwidth}{@{\extracolsep{\fill}}lcccc@{}}
\toprule
\multicolumn{5}{l}{\textit{(a) Mean recorded-action $L_2$ error (action source / $q_t$)}}  \\
Policy
& Demo / $\mathtt{ref}$
& Demo / $\mathtt{quant}$
& FP / $\mathtt{ref}$
& FP / $\mathtt{quant}$ \\
\midrule
PTQ
& $1.852\pm0.731$
& $1.852\pm0.731$
& $1.969\pm1.056$
& $1.969\pm1.056$ \\
QCRL
& $1.265\pm0.675$
& $1.305\pm0.658$
& $1.456\pm0.878$
& $1.495\pm0.913$ \\
\bottomrule
\end{tabular*}
\par\vspace{3pt}
\begin{tabular*}{\columnwidth}{@{\extracolsep{\fill}}lccc@{}}
\toprule
\multicolumn{4}{l}{\textit{(b) Collected rollouts: full success/progress (\%)}} \\
Source & Folding & Clean desk & Short tasks \\
\midrule
FP & 36.7/58.3 & 46.7/69.2 & 86.0/90.0 \\
PTQ & 0.0/13.3 & 16.7/39.2 & 69.3/76.7 \\
\bottomrule
\end{tabular*}
\par\vspace{3pt}
\begin{tabular*}{\columnwidth}{@{\extracolsep{\fill}}lcc@{}}
\toprule
\multicolumn{3}{l}{\textit{(c) $L_2$ distance in normalized action space}} \\
Source & To PTQ & Within source \\
\midrule
FP counterpart & $0.87\pm0.21$ & $1.06\pm0.39$ \\
External $\pi_0$ & $18.07\pm3.57$ & $11.41\pm4.11$ \\
\bottomrule
\end{tabular*}
\end{table}
\subsection{Analysis}
\label{sec:source_selection_analysis}
\begin{table*}[!t]
\centering
\caption{QCRL data, evaluation, and training settings. Collected episodes: Demo/FP/PTQ; batch sizes: global/micro.}
\vspace{-0.3cm}
\label{tab:reproducibility_pending}
\footnotesize
\setlength{\tabcolsep}{3pt}
\renewcommand{\arraystretch}{1.08}
\begin{minipage}[t]{0.485\textwidth}
\vspace{0pt}
\textit{(a) Data, evaluation, and conditioning}\par\smallskip
\begin{tabularx}{\linewidth}{@{}l>{\raggedright\arraybackslash}X@{}}
\toprule
Item & Setting \\
\midrule
Real-world data & Per task: short 150/50/50; long 80/30/30 \\
Task grouping & One policy for three short tasks; one each for folding and clean desk \\
LIBERO data & Per model: four suites (40 tasks),
1,693/2,000/2,000 (5,693 total) \\
 & FP/PTQ: 500 per suite each; seeds 42/7 \\
\midrule
Evaluation trials & Real: short ID 30, OOD/long 10; LIBERO: 500 per task \\
Episode limits & LIBERO Spatial/Object/Goal: 480; Long: 520 steps \\
\midrule
RL / guidance & $\gamma=1$, $N=8$, $\beta_u=3/7$; guidance scale 1 \\
Penalty $\lambda_f$ & LIBERO: 300; real short/long: 600/1500 \\
\bottomrule
\end{tabularx}
\end{minipage}\hfill
\begin{minipage}[t]{0.495\textwidth}
\vspace{0pt}
\textit{(b) Training and optimization defaults}\par\smallskip
\begin{tabularx}{\linewidth}{@{}l>{\raggedright\arraybackslash}X>{\raggedright\arraybackslash}X@{}}
\toprule
Setting & Value model & Policy \\
\midrule
Updates / warmup & 18k / 500 & 30k / 5k \\
Learning rate & Backbone: $5\!\times\!10^{-5}$; \newline head: $10^{-4}$ & $10^{-5}$ \\
Schedule & Constant & Cosine to zero \\
Batch (LIBERO) & 256/32 & 128/16 \\
Batch (real-world) & 128/16 & 128/16 \\
\midrule
\end{tabularx}
\par
\begin{tabularx}{\linewidth}{@{}l>{\raggedright\arraybackslash}X@{}}
Optimizer (both) & AdamW; $\beta=(0.9,0.95)$, $\epsilon=10^{-8}$ \\
Source sampling & Demo:FP:PTQ = 1:1:1 \\
Decay / clipping & Weight decay $10^{-10}$; gradient clipping 1.0 \\
Value model & SigLIP2 + Gemma-3-270M; 201-bin categorical head;
training follows RECAP~\cite{amin2025pi06} \\ 
\bottomrule
\end{tabularx}
\end{minipage}
\vspace{-0.3cm}
\end{table*}
\newparagraph{Setting}
We examine when approximating positive reference behavior can improve on $\pi_Q^0$. Consider a horizon-$T$ Markov process with discount $1$, a shared initial distribution, and zero-reward padding after termination. The state includes the instruction, sufficient history, and milestone-completion status. Let $J_T$ denote expected return, $d_t^0$ the PTQ state distribution, and $A_t^0=Q_t^{\pi_Q^0}-V_t^{\pi_Q^0}$ its exact advantage, with $|A_t^0|\leq A_{\max}$. For an idealized positive-reference policy $\pi_r^+$ defined at every state, let
\begin{equation}
\label{eq:source_advantage_tv}
\begin{aligned}
S_r&=\sum_{t=1}^T\mathbb E_{s\sim d_t^0,\,a\sim\pi_{r,t}^+}[A_t^0(s,a)],\\
\epsilon_{\mathrm{BC}}&=\sup_{t,s}\mathrm{TV}(\pi_{\mathrm{deploy},t}(\cdot\mid s),\pi_{r,t}^+(\cdot\mid s)),\\
\alpha_r&=\sup_{t,s}\mathrm{TV}(\pi_{r,t}^+(\cdot\mid s),\pi_{Q,t}^0(\cdot\mid s)).
\end{aligned}
\end{equation}

TV is half the L1 distance between action distributions. Here, $S_r$ measures reference-action benefit at PTQ states, $\epsilon_{\mathrm{BC}}$ measures approximation error, and $\alpha_r$ measures reference--PTQ mismatch. The performance-difference identity~\cite{kakade2002approximately,schulman2015trpo} and finite-horizon TV bounds yield
\begin{equation}
\label{eq:deployment_policy_bound}
\begin{aligned}
J_T(\pi_{\mathrm{deploy}})-J_T(\pi_Q^0)
&\geq S_r-2TA_{\max}\epsilon_{\mathrm{BC}}\\
&\quad-2A_{\max}T(T-1)(\alpha_r+\epsilon_{\mathrm{BC}})^2.
\end{aligned}
\end{equation}

The bound is positive when reference-action benefit exceeds approximation and state-distribution penalties, with the latter growing with the horizon. This links FP source selection to useful behavior near PTQ states and conditioning to reproducing the selected reference behavior. We use this as a design guide: training uses fitted advantages, and the diagnostics below do not estimate the worst-case TV terms.

\newparagraph{Empirical support}
Table~\ref{tab:source_diagnostics} supports the design through three diagnostics. (a) QCRL reduces mean $L_2$ error against the recorded 50-step action targets from demonstrations and FP rollouts, with lower means under $q_t=\mathtt{ref}$ than $q_t=\mathtt{quant}$, supporting reference-behavior selection. (b) Collected FP rollouts achieve higher success and progress than PTQ rollouts, providing stronger reference experience. (c) On 100 shared PTQ observations with matched Gaussian noise, the FP counterpart has a smaller $L_2$ distance to PTQ
in normalized action space than a separately trained $\pi_0$ ($0.87$ versus $18.07$); independent noise measures within-source variation. Together, these observations support learning from higher-quality reference behavior with limited observed action deviation. Section~\ref{sec:qcrl_ablation_results} separately evaluates the task-performance effects of FP rollout augmentation, a progress-scaled terminal penalty, and quantization-aware conditioning.

\section{Experiments}
\label{sec:experiments}

We evaluate QCRL under two deployment settings: INT4 W4A16 on $\pi_{0.5}$ and GR00T N1.5, and NVFP4 W4A4 on $\pi_{0.5}$. We assess task recovery, OOD generalization, component contributions, and hardware efficiency.

\subsection{Experimental Setup}
\label{sec:experimental_setup}

\newparagraph{Precision settings}
INT4 W4A16 follows Section~\ref{sec:comparative_protocol}. NVFP4 W4A4 quantizes language-model weights and activations, with FP8 vision and action modules. PTQ, QAT, and QCRL use the same runtime at each precision, and each deploys a single quantized policy. TensorRT measurements use FP16 as the baseline precision on Jetson Thor and B200; NVFP4 runtime results use 1,000 samples.

\newparagraph{Data and evaluation}
We use four LIBERO suites~\cite{libero} and the real-world tasks in Fig.~\ref{fig:task_overview}. Table~\ref{tab:reproducibility_pending}(a) specifies data composition, task grouping, and evaluation budgets; Table~\ref{tab:id_task_instructions} lists task instructions. Real-world progress follows Section~III-B; LIBERO computes
$p(\tau)$ at termination using task-ordered simulator checks
for object--gripper contact, height gain exceeding $0.10$\,m from the object's initial height, and target-region containment.
FP augmentation uses the existing checkpoint without new expert demonstrations. All adaptation uses ID data; visual and semantic shifts test OOD transfer using the metrics in Section~\ref{sec:task_protocol}.

\begin{table}[!htbp]
\centering
\caption{Real-world ID task instructions.}
\vspace{-0.2cm}
\label{tab:id_task_instructions}
\footnotesize
\setlength{\tabcolsep}{3pt}
\renewcommand{\arraystretch}{1.08}
\begin{tabularx}{\columnwidth}{@{}l>{\raggedright\arraybackslash}X@{}}
\toprule
Task & Instruction \\
\midrule
Sweep & sweep the gray cloth to the right side of the table \\
Pick/place & pick up the corn and place it on the plate \\
Stack & pick up the pink cup and stack it inside the blue cup \\
Folding & fold the blue cloth in half twice \\
Clean desk & clean the desk by putting items into the green basket and trash into the yellow box \\
\bottomrule
\end{tabularx}
\end{table}

\newparagraph{Baselines}
PTQ quantizes the task-trained FP baseline; SQIL QAT and
QCRL initialize from the FP checkpoint. RECAP+QAT~\cite{amin2025pi06} uses demonstrations and quantized rollouts with the same 30k policy-update budget as QCRL, serving as the baseline without QCRL’s three added components. For LIBERO, we additionally include FP RECAP; for NVFP4, we include an FP QCRL control with the same rollouts and update budget to distinguish general adaptation from low-bit recovery.

\newparagraph{Training and advantage labeling}
\label{sec:implementation_settings}
We fit the value model for 18k updates, compute advantage labels, and train the quantized policy for 30k updates using the source sampling and optimizer settings in Table~\ref{tab:reproducibility_pending}(b). Both $\pi_{0.5}$ and GR00T N1.5 use flow-matching action losses~\cite{pi05,grootn15modelcard}. At termination, including time limits, advantage targets use rewards through the final step and zero bootstrap. Following RECAP~\cite{amin2025pi06}, we set each task's threshold $\epsilon_\ell$ to the 30th percentile of estimated advantages for that task; ties are labeled positive. The value checkpoint and threshold remain fixed during policy training. The conditioning labels are \{\texttt{positive}, \texttt{negative}\} and \{\texttt{ref}, \texttt{quant}\}; inference fixes the positive reference condition with guidance scale 1.

\begin{figure}[!t]
\centering
\includegraphics[width=0.96\columnwidth]{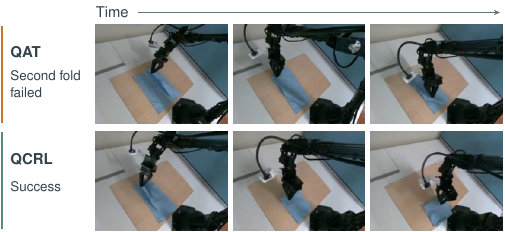}
\vspace{-0.1cm}
\caption{Cloth folding ($\pi_{0.5}$, INT4 W4A16): QAT fails at the second fold; QCRL completes it. Additional tasks are shown in the supplementary video.}
\label{fig:rollout_examples}
\vspace{-0.2cm}
\end{figure}

\subsection{Task Performance and OOD Generalization}
\label{sec:main_recovery_results}
\label{sec:cross_format_evaluation}

\newparagraph{Real-world ID performance}
QCRL improves long-horizon execution while maintaining or improving short-task full success rates over QAT (Table~\ref{tab:realworld_id_results}). For $\pi_{0.5}$, it completes folding where QAT achieves only partial progress (Fig.~\ref{fig:rollout_examples}) and restores the clean-desk full success rate to the FP level. For GR00T, QCRL restores the folding full success rate to the FP level, but its gains on clean desk remain limited to partial completion.

\begin{table}[!t]
\centering
\caption{Real-world ID performance (\%; quantized: INT4 W4A16).
Full success unless marked prog. (progress)}
\vspace{-0.2cm}
\label{tab:realworld_id_results}
\footnotesize
\setlength{\tabcolsep}{1pt}
\renewcommand{\arraystretch}{1.08}
\begin{tabular*}{\columnwidth}{@{\extracolsep{\fill}}l*{8}{N{3.1}}@{}}
\toprule
\multirow{2}{*}{Task}
& \multicolumn{4}{c}{$\pi_{0.5}$}
& \multicolumn{4}{c}{GR00T N1.5} \\
\cmidrule(lr){2-5}\cmidrule(lr){6-9}
& \multicolumn{1}{c}{FP}
& \multicolumn{1}{c}{PTQ}
& \multicolumn{1}{c}{QAT}
& \multicolumn{1}{c}{QCRL}
& \multicolumn{1}{c}{FP}
& \multicolumn{1}{c}{PTQ}
& \multicolumn{1}{c}{QAT}
& \multicolumn{1}{c}{QCRL} \\
\midrule
Sweep
& 100.0 & 100.0 & 100.0 & 100.0
& 100.0 & 93.3 & 96.7 & 100.0 \\
Pick/place
& 76.7 & 63.3 & 76.7 & 80.0
& 50.0 & 30.0 & 40.0 & 40.0 \\
Stack
& 83.3 & 46.7 & 83.3 & 83.3
& 20.0 & 10.0 & 13.3 & 23.3 \\
\midrule
Folding (full)
& 40.0 & 0.0 & 0.0 & 30.0
& 40.0 & 30.0 & 30.0 & 40.0 \\
Folding (prog.)
& 60.0 & 15.0 & 40.0 & 60.0
& 55.0 & 40.0 & 45.0 & 60.0 \\
Clean desk (full)
& 50.0 & 20.0 & 30.0 & 50.0
& 0.0 & 0.0 & 0.0 & 0.0 \\
Clean desk (prog.)
& 70.0 & 40.0 & 52.5 & 65.0
& 25.0 & 12.5 & 15.0 & 22.5 \\
\bottomrule
\end{tabular*}
\vspace{-0.1cm}
\end{table}

\begingroup
\clubpenalties 2 10000 0\relax
\newparagraph{OOD generalization}
In W4A16, ID-only QCRL adaptation narrows $\pi_{0.5}$'s residual QAT gaps under sensor degradation and unseen-object shifts, while matching FP under appearance and illumination changes (Table~\ref{tab:realworld_qcrl_results}). GR00T shows a similar trend, exceeding FP on illumination and unseen objects.
\par
\endgroup

NVFP4 likewise shows PTQ degradation, incomplete QAT recovery, and further QCRL gains (Table~\ref{tab:nvfp4_results}). Under appearance shifts, $\pi_{0.5}$ improves from 60.0\% with QAT to 83.3\% with QCRL; under added context, it improves from 45.0\% to 75.0\%. The relative difficulty of shifts differs between formats. Section~\ref{sec:adaptation_quantization_recovery} distinguishes general adaptation from low-bit recovery.

\begin{table}[!t]
\centering
\caption{OOD and hardware results (quantized: INT4 W4A16).
ID-only adaptation. Visual/semantic scores (\%): task score/primitive progress.}
\vspace{-0.2cm}
\label{tab:realworld_qcrl_results}
\label{tab:hardware_w4a16}
\footnotesize
\setlength{\tabcolsep}{2pt}
\renewcommand{\arraystretch}{1.08}
\begin{tabular*}{\columnwidth}{@{\extracolsep{\fill}}l*{8}{N{2.1}}@{}}
\toprule
\multirow{2}{*}{Condition} & \multicolumn{4}{c}{$\pi_{0.5}$} & \multicolumn{4}{c}{GR00T N1.5} \\
\cmidrule(lr){2-5}\cmidrule(lr){6-9}
 & \multicolumn{1}{c}{FP} & \multicolumn{1}{c}{PTQ} & \multicolumn{1}{c}{QAT} & \multicolumn{1}{c}{QCRL} & \multicolumn{1}{c}{FP} & \multicolumn{1}{c}{PTQ} & \multicolumn{1}{c}{QAT} & \multicolumn{1}{c}{QCRL} \\
\midrule
Appearance & 86.7 & 73.3 & 83.3 & 86.7 & 51.7 & 41.7 & 50.0 & 55.0 \\
Illumination & 73.3 & 46.7 & 53.3 & 73.3 & 63.3 & 55.0 & 55.0 & 86.7 \\
Sensor degradation & 70.0 & 13.3 & 41.7 & 68.3 & 45.0 & 21.7 & 20.0 & 45.0 \\
\midrule
Unseen object & 87.5 & 45.0 & 67.5 & 85.0 & 52.5 & 42.5 & 45.0 & 67.5 \\
Role reversal & 43.3 & 43.3 & 40.0 & 60.0 & 33.3 & 33.3 & 30.0 & 40.0 \\
\bottomrule
\end{tabular*}

\par\vspace{3pt}
\begin{tabular*}{\columnwidth}
{@{\extracolsep{\fill}}l*{2}{N{1.2}N{1.2}N{3.2}}@{}}
\toprule
\multicolumn{7}{l}
{\textit{$\pi_{0.5}$ TensorRT (memory: GiB; power: W)}} \\
\multirow{2}{*}{Format}
& \multicolumn{3}{c}{Jetson Thor}
& \multicolumn{3}{c}{B200} \\
\cmidrule(lr){2-4}\cmidrule(lr){5-7}
& \multicolumn{1}{c}{Model}
& \multicolumn{1}{c}{Peak}
& \multicolumn{1}{c}{Power}
& \multicolumn{1}{c}{Model}
& \multicolumn{1}{c}{Peak}
& \multicolumn{1}{c}{Power} \\
\midrule
FP16
& 6.46 & 8.02 & 105.80
& 6.46 & 6.49 & 584.92 \\
INT4 W4A16
& 3.69 & 5.31 & 105.04
& 3.69 & 4.13 & 550.18 \\
\bottomrule
\end{tabular*}
\vspace{-0.3cm}
\end{table}

\begin{table}[!t]
\centering
\caption{Task and hardware results for $\pi_{0.5}$ NVFP4 W4A4. Scores (\%): simulation success, visual task score, semantic primitive progress. FP Base/Adapt.: task-trained baseline/FP QCRL control.}
\vspace{-0.2cm}
\label{tab:nvfp4_results}
\label{tab:hardware_nvfp4}
\label{tab:hardware_efficiency}
\footnotesize
\setlength{\tabcolsep}{2pt}
\renewcommand{\arraystretch}{1.1}
\begin{tabular*}{\columnwidth}{@{\extracolsep{\fill}}llN{2.1}N{2.1}N{2.1}N{2.1}N{2.1}@{}}
\toprule
\multirow{2}{*}{Type}
& \multirow{2}{*}{Condition}
& \multicolumn{2}{c}{FP}
& \multicolumn{3}{c}{NVFP4 W4A4} \\
\cmidrule(lr){3-4}
\cmidrule(lr){5-7}
& &
\multicolumn{1}{c}{Base}
& \multicolumn{1}{c}{Adapt.}
& \multicolumn{1}{c}{PTQ}
& \multicolumn{1}{c}{QAT}
& \multicolumn{1}{c}{QCRL} \\
\midrule
Simulation
& LIBERO-Long
& 93.2 & 94.1 & 92.5 & 93.0 & 94.0 \\
\midrule
\multirow{3}{*}{Visual}
& Appearance
& 86.7 & 86.7 & 48.3 & 60.0 & 83.3 \\
& Illumination
& 73.3 & 70.0 & 40.0 & 70.0 & 73.3 \\
& Sensor degradation
& 70.0 & 73.3 & 43.3 & 70.0 & 75.0 \\
\cmidrule(lr){2-7}
\multirow{3}{*}{Semantic}
& Unseen object
& 87.5 & 87.5 & 75.0 & 80.0 & 80.0 \\
& Role reversal
& 43.3 & 63.3 & 43.3 & 50.0 & 70.0 \\
& Added context
& 70.0 & 77.5 & 37.5 & 45.0 & 75.0 \\
\bottomrule
\end{tabular*}

\par\vspace{3pt}
\begin{tabular*}{\columnwidth}{@{\extracolsep{\fill}}llN{1.2}N{1.2}cN{3.2}@{}}
\toprule
\multicolumn{6}{l}{\textit{$\pi_{0.5}$ TensorRT: FP16 vs. NVFP4/FP8}} \\
Device & Format & \multicolumn{1}{c}{\makecell{Model\\(GiB)}} & \multicolumn{1}{c}{\makecell{Peak\\(GiB)}} & \multicolumn{1}{c}{\makecell{Latency\\(ms)}} & \multicolumn{1}{c}{\makecell{Power\\(W)}} \\
\midrule
\multirow{2}{*}{Jetson Thor} & FP16 & 6.46 & 8.02 & $96.61\pm2.37$ & 105.80 \\
& NVFP4 & 2.80 & 2.81 & $55.85\pm0.10$ & 89.11 \\
\addlinespace[2pt]
\multirow{2}{*}{B200} & FP16 & 6.46 & 6.49 & $17.68\pm0.12$ & 584.92 \\
& NVFP4 & 2.80 & 3.30 & $14.16\pm0.10$ & 439.76 \\
\bottomrule
\end{tabular*}
\end{table}
\begingroup
\clubpenalty=10000
\newparagraph{LIBERO performance}
Under W4A16, QCRL maintains $\pi_{0.5}$'s near-saturated Spatial, Object, and Goal performance, with clearer gains over PTQ and RECAP+QAT on LIBERO-Long (Table~\ref{tab:libero_qcrl_results}). For GR00T, QCRL matches or exceeds FP across all four suites, supporting recovery in sequential execution while retaining performance on shorter benchmarks.\par
\endgroup

\begin{table}[!t]
\centering
\caption{LIBERO full success (\%; quantized methods: INT4 W4A16).}
\vspace{-0.2cm}
\label{tab:libero_qcrl_results}
\footnotesize
\setlength{\tabcolsep}{2pt}
\renewcommand{\arraystretch}{1.08}
\begin{tabular*}{\columnwidth}{@{\extracolsep{\fill}}l*{8}{N{2.1}}@{}}
\toprule
\multirow{2}{*}{Method} & \multicolumn{4}{c}{$\pi_{0.5}$} & \multicolumn{4}{c}{GR00T N1.5} \\
\cmidrule(lr){2-5}\cmidrule(lr){6-9}
 & \multicolumn{1}{c}{Spatial} & \multicolumn{1}{c}{Object} & \multicolumn{1}{c}{Goal} & \multicolumn{1}{c}{Long} & \multicolumn{1}{c}{Spatial} & \multicolumn{1}{c}{Object} & \multicolumn{1}{c}{Goal} & \multicolumn{1}{c}{Long} \\
\midrule
FP & 99.0 & 98.8 & 98.2 & 93.2 & 96.8 & 98.4 & 95.4 & 87.0 \\
FP RECAP & 99.8 & 99.0 & 98.6 & 94.4 & 96.8 & 98.2 & 95.6 & 87.4 \\
\midrule
PTQ & 98.8 & 98.4 & 97.8 & 92.4 & 96.2 & 97.4 & 92.0 & 85.2 \\
RECAP+QAT & 99.2 & 99.0 & 98.5 & 93.0 & 96.2 & 97.6 & 92.2 & 85.8 \\
QCRL & 99.4 & 99.2 & 98.4 & 94.8 & 96.8 & 98.6 & 95.8 & 87.8 \\
\bottomrule\end{tabular*}
\vspace{-0.3cm}
\end{table}

Under NVFP4, QCRL raises $\pi_{0.5}$ LIBERO-Long success from 93.0\% with QAT to 94.0\%, approaching the adapted FP control at 94.1\% (Table~\ref{tab:nvfp4_results}). Both formats therefore support improved sequential execution, although the absolute gains on this near-saturated suite are modest.

\subsection{Adaptation and Quantization Recovery}
\label{sec:adaptation_quantization_recovery}

On LIBERO-Long, QCRL outperforms RECAP+QAT for both models
under the same policy-update budget and modestly exceeds FP
RECAP while retaining quantized inference.

Using the FP control in Table~\ref{tab:nvfp4_results}, we compare
adaptation gains, $\Delta_{\mathrm{FP}}=\mathrm{FP~Adapt.}-\mathrm{FP~Base}$
and $\Delta_{\mathrm{Q}}=\mathrm{QCRL}-\mathrm{QAT}$.
Under appearance shifts, $\Delta_{\mathrm{FP}}=0.0$ percentage points while
$\Delta_{\mathrm{Q}}=23.3$ percentage points. With added context, the gains are
7.5 and 30.0 percentage points, respectively. These conditions are consistent
with low-bit recovery beyond the corresponding FP adaptation gains.

Recovery is condition-dependent. Role reversal yields equal 20.0-point
primitive-progress gains at both precisions, while unseen-object performance
is unchanged by adaptation in both cases. QCRL therefore
combines general policy improvement with condition-dependent reductions
in low-bit gaps. Differences in starting performance and ceiling effects
prevent attributing all gains exclusively to quantization-specific recovery.

\subsection{Component and Conditioning Ablation}
\label{sec:qcrl_ablation_results}
Table~\ref{tab:qcrl_ablation} compares all eight combinations using the same demonstrations, PTQ rollouts, and update budgets. The all-off baseline is RECAP+QAT with rollout-based adaptation (Sensor: 58.3\%), distinct from standard SQIL QAT in Table~\ref{tab:realworld_qcrl_results} (41.7\%). Variants with quantization-aware conditioning use $q_t=\mathtt{ref}$ at evaluation; without FP rollouts, demonstrations alone provide the reference data.

Progress scaling improves sensor-degradation task scores for every combination of the other components, while LIBERO-Long success improves or remains unchanged. FP rollout augmentation yields the largest improvement under sensor degradation among individual components. QCRL achieves the highest scores, although these point estimates do not establish the statistical significance of small differences.

The lower panel of Table~\ref{tab:qcrl_ablation} holds the QCRL checkpoint, quantization, and $I_t=\mathrm{pos}$ fixed. The reference condition yields higher scores on both evaluations. Token removal also lowers scores but changes the conditioning pattern, making it a separate sensitivity test.

\begin{table}[!t]
\centering
\vspace{-0.2cm}
\caption{QCRL ablations ($\pi_{0.5}$, INT4 W4A16; \%). All-off/on: RECAP+QAT/QCRL. Prec. cond.: quantization-aware conditioning. Long/Sensor: LIBERO-Long success/sensor-degradation task score.}
\vspace{-0.2cm}
\label{tab:qcrl_ablation}
\footnotesize
\setlength{\tabcolsep}{2pt}
\renewcommand{\arraystretch}{1.08}
\begin{tabular*}{\columnwidth}{@{\extracolsep{\fill}}l*{8}{c}@{}}
\toprule
FP roll.   & -- & \checkmark & -- & -- & \checkmark & \checkmark & -- & \checkmark \\
Prog.      & -- & -- & \checkmark & -- & \checkmark & -- & \checkmark & \checkmark \\
Prec. cond. & -- & -- & -- & \checkmark & -- & \checkmark & \checkmark & \checkmark \\
\midrule
Long   & 93.0 & 94.0 & 93.4 & 93.2 & 94.0 & 94.4 & 93.8 & \textbf{94.8} \\
Sensor & 58.3 & 65.0 & 60.0 & 60.0 & 66.7 & 66.7 & 63.3 & \textbf{68.3} \\
\bottomrule
\end{tabular*}
\par\vspace{3pt}
\begin{tabular*}{\columnwidth}{@{\extracolsep{\fill}}lccc@{}}
\toprule
Deployment
& $q_t=\mathtt{ref}$
& $q_t$ removed
& $q_t=\mathtt{quant}$ \\
\midrule
Long / Sensor
& \textbf{94.8 / 68.3}
& 94.0 / 65.0
& 93.8 / 65.0 \\
\bottomrule
\end{tabular*}
\vspace{-0.4cm}
\end{table}

\subsection{Deployment Efficiency}
\label{sec:deployment_cost}

INT4 W4A16 reduces model and peak memory relative to FP16 on both devices (Table~\ref{tab:hardware_w4a16}). The NVFP4/FP8 configuration reduces model memory, peak memory, latency, and power (Table~\ref{tab:hardware_nvfp4}); on Jetson Thor, latency decreases from 96.61 to 55.85\,ms. These runtime benefits depend on precision and are shared by PTQ, QAT, and QCRL.

\section{Conclusion}
\label{sec:conclusion}

QCRL combines FP rollout augmentation, a progress-scaled terminal penalty, and quantization-aware conditioning to address residual quantized VLA failures. Simulation and real-world experiments show improved long-horizon execution and OOD generalization. NVFP4 deployment of $\pi_{0.5}$ combines this recovery with reduced memory use and latency on Jetson Thor and B200.

\bibliographystyle{IEEEtran}
\bibliography{references}

\clearpage

\useRomanappendicesfalse
\appendices
\appendixfloatlayout
\setcounter{table}{0}
\renewcommand{\thetable}{A\arabic{table}}
\renewcommand{\theHtable}{appendix.\arabic{table}}
\setcounter{equation}{0}
\renewcommand{\theequation}{A\arabic{equation}}
\renewcommand{\theHequation}{appendix.\arabic{equation}}
\renewcommand{\thesubsection}{\thesection.\arabic{subsection}}
\renewcommand{\thesubsectiondis}{\thesubsection.}
\section{Supplementary Experimental Details and Results}
\label{app:supplementary}
\subsection{Experimental Protocol and Implementation Details}
\label{app:experimental_protocol_details}

This appendix expands the evaluation protocol and provides the task-level evidence behind the main paper. The advantage-conditioned objective is given in Section~\ref{sec:recap_preliminaries}; the robot setup, baseline schedules, and QCRL data and optimization settings are given in Sections~\ref{sec:comparative_protocol} and~\ref{sec:experimental_setup} and Tables~\ref{tab:baseline_training} and~\ref{tab:reproducibility_pending}. We refer to those descriptions for settings already specified in the main text.

\paragraph{Data collection and task grouping.}
The same Trossen ViperX tabletop platform and base/wrist camera setup are used for supervised demonstration collection, FP and quantized rollout collection, and real-world evaluation. The three short-horizon tasks are combined into one adapted policy because they share reaching, grasping, moving, placing, and releasing primitives. Cloth folding and clean desk use separate policies because their ordered subgoal structures differ. The fixed mixture contains demonstrations and autonomous rollouts, including unsuccessful and partially completed episodes. No visual-shift or semantic-shift observations are added during adaptation; those settings are used only to evaluate transfer from ID training.

\paragraph{Visual shifts.}
Appearance changes replace the cloth color in sweeping and the background in pick-and-place and stacking. Illumination changes reduce image brightness/contrast, and sensor degradation blurs observations. Distractor settings introduce additional task-irrelevant objects. The first three shift families are evaluated on all three short tasks; distractors are evaluated on pick-and-place and stacking. These choices explain the three-task versus two-task averages in the main paper. Table~\ref{app:tab:ood_protocol} summarizes the additional evaluation conditions; representative observations appear in Fig.~\ref{fig:task_overview}.

\paragraph{Object and instruction shifts.}
The pick-and-place evaluation separately replaces the manipulated object (corn with carrot) and the recipient (plate with pot). Role reversal exchanges the instructed roles of the pink and blue cups. Added-context prompts modify the language input while preserving the target manipulation: the pick-and-place prompt adds a preliminary action and the stacking prompt adds task-focus context. Their exact wording is given in Table~\ref{app:tab:shift_prompts}; the five original ID prompts are already listed in Table~\ref{tab:id_task_instructions}.

\begin{table*}[t]
\centering
\caption{Additional OOD evaluation definitions. Each evaluated task--condition pair uses 10 trials per model and method. The main paper defines full success and milestone progress.}
\label{app:tab:ood_protocol}
\footnotesize
\setlength{\tabcolsep}{4pt}
\renewcommand{\arraystretch}{1.12}
\begin{tabularx}{\textwidth}{@{}ll>{\raggedright\arraybackslash}X>{\raggedright\arraybackslash}X@{}}
\toprule
Group & Condition & Change from the ID setting & Interpretation of the reported score \\
\midrule
Visual & Appearance & Unseen cloth color or unseen background. & Sweeping success; pick/place or stacking progress. \\
 & Illumination & Reduced image brightness/contrast. & Same task metrics under changed observations. \\
 & Sensor degradation & Blurred observations. & Same task metrics under degraded observations. \\
 & Distractors & Additional task-irrelevant objects. & Pick/place or stacking progress. \\
\midrule
Semantic & Object / recipient & Corn $\rightarrow$ carrot; plate $\rightarrow$ pot. & Mean completion of pick and place stages. \\
 & Role reversal & Reverse the pink and blue cups' instructed roles. & Main-paper primitive progress averages move-to-target, pick, and place, without certifying correct roles. \\
 & Added context & Preliminary-action or task-focus language; see Table~\ref{app:tab:shift_prompts}. & Progress on the specified manipulation. \\
 & Null input & Literal string \texttt{null}. & Task-like execution without a valid task instruction; a diagnostic, not instruction-following success. \\
\bottomrule
\end{tabularx}
\end{table*}

\begin{table*}[t]
\centering
\caption{Added-context and null-input prompts used only for semantic evaluation.}
\label{app:tab:shift_prompts}
\footnotesize
\setlength{\tabcolsep}{4pt}
\renewcommand{\arraystretch}{1.12}
\begin{tabularx}{\textwidth}{@{}ll>{\raggedright\arraybackslash}X@{}}
\toprule
Task & Condition & Input instruction \\
\midrule
Pick-and-place & Added context & before doing anything else, look at the carrot, then pick up the corn and place it on the plate \\
Pick-and-place & Null input & \texttt{null} \\
Stacking & Added context & you only need to focus on the cups. pick up the pink cup and stack it inside the blue cup \\
Stacking & Null input & \texttt{null} \\
\bottomrule
\end{tabularx}
\end{table*}

\paragraph{Reading semantic metrics.}
Primitive progress measures execution of task-specific manipulation stages and need not imply correct selection of the instruction-specified object or role. Likewise, the added-context scores do not separately certify every language clause. Null-input scores record task-like manipulation despite the absence of a valid task instruction; higher scores are not better instruction following, and lower scores alone do not establish a learned abstention policy. We retain these diagnostics separately from the valid-instruction results.

\paragraph{Progress and offline learning.}
Final partial completion affects only the trajectory's terminal penalty, rather than supplying a dense reward at each milestone. The value model is fitted before policy training, and its fixed estimates determine the transition-level advantage labels. Consequently, partial trajectory success does not make every transition positive. The source condition identifies data provenance rather than arithmetic precision: demonstrations and FP rollouts form the reference group, and the deployed policy remains quantized when selecting reference behavior. All value learning, return computation, and source annotation take place offline.

\subsection{Low-Bit Quantization and Additional Training Configuration}
\label{app:quantization_details}

The comparative tables in this appendix concern INT4 W4A16, with group-wise asymmetric four-bit language-model weights and group size 128. The vision and action modules retain their original precision. This scope differs from the NVFP4/FP8 experiments in the main paper; the latter are not represented by the additional INT4 tables below. The general quantizer and deployment-format definitions are given in Section~\ref{sec:related_work}.

PTQ applies weight quantization after task training. QAT simulates quantized weights during adaptation so that trainable parameters can accommodate the low-bit representation; the evaluated SQIL objective additionally uses saliency-guided FP-to-quantized distillation~\cite{park2025sqil}. QCRL retains the quantized policy while using rollout returns and advantage labels for policy improvement. Thus, the extra adaptation changes how the policy is trained, while the runtime precision remains shared across methods within each format.

\paragraph{Longer-trained SmolVLA.}
The notation \textit{SmolVLA-L} denotes an additional configuration with the same architecture and parameter count as SmolVLA. Its FP training uses 32k steps at learning rate $2\times10^{-4}$, and its QAT run uses 32k steps at $2\times10^{-5}$, both with batch size 64 in LeRobot. PTQ quantizes that FP checkpoint. Relative to the main-paper SmolVLA baseline, this is $10\times$ the FP training steps and $20\times$ the QAT steps, with a different QAT learning rate. It is therefore an auxiliary training configuration rather than a controlled, matched-budget ablation. Its nonduplicated results are retained in Table~\ref{app:tab:smolvla_longer_training}.

\subsection{Full Comparative Result Tables}
\label{app:full_comparative_results}

Table~\ref{tab:visual_generalization_summary} reports the visual-condition summary across all five model families, while Table~\ref{tab:visual_ood_robustness} provides the complete per-task and per-model breakdown. The latter table uses one row per precision/recovery method, showing each FP, PTQ, and QAT comparison separately. All scores are percentages. Sweeping uses full success; pick-and-place and stacking use mean completion of their ordered milestones.

Appearance, illumination, and sensor-degradation means each cover three tasks and five models. Distractor means cover the two evaluated tasks and five models. Consequently, a mean over the four condition means is not the same as a mean over all 11 task--condition pairs. Parenthetical values are percentage-point drops from FP, computed before rounding, rather than relative percentage losses.

The five main models' short-task full success and long-task full/progress scores are already reported in Table~\ref{tab:comparative_models}; those identical rows are not repeated. Table~\ref{app:tab:smolvla_longer_training} reports the additional SmolVLA-L results. Longer training improves several short-task scores, but neither FP nor its low-bit variants completes the full folding or clean-desk task in these evaluations. The different training budgets and learning rates should be kept in mind when comparing this configuration to the main baseline.

\begin{table*}[t]
\centering
\caption{
Visual generalization task-score summary across the four visual shifts. Scores are averaged over evaluated task and model entries. For cloth sweep the score denotes full-task success, and for pick-and-place and cup stacking it denotes the ordered-milestone task score. All quantized methods use INT4 W4A16. Each cell reports score with drop from FP in parentheses ($\Delta = \mathrm{FP} - \mathrm{Method}$); a smaller parenthetical means closer to FP.
}
\label{tab:visual_generalization_summary}

\footnotesize
\setlength{\tabcolsep}{4pt}
\renewcommand{\arraystretch}{1.08}

\begin{tabular*}{\textwidth}{@{\extracolsep{\fill}}llccc@{}}
\toprule
Visual condition & Included settings
& FP
& PTQ
& QAT \\
\midrule

Appearance shift
& Unseen color / unseen background
& 60.3 {\scriptsize (0.0)}
& 48.0 {\scriptsize (12.3)}
& 56.3 {\scriptsize (4.0)} \\

Illumination shift
& Brightness / contrast shift
& 61.7 {\scriptsize (0.0)}
& 47.0 {\scriptsize (14.7)}
& 51.7 {\scriptsize (10.0)} \\

Sensor degradation
& Blurred observation
& 48.0 {\scriptsize (0.0)}
& 19.0 {\scriptsize (29.0)}
& 28.7 {\scriptsize (19.3)} \\

Distractor
& Additional irrelevant objects
& 16.0 {\scriptsize (0.0)}
& 8.5 {\scriptsize (7.5)}
& 10.5 {\scriptsize (5.5)} \\

\bottomrule
\end{tabular*}
\end{table*}

\begin{table*}[t]
\centering
\begin{minipage}{0.78\textwidth}
\centering
\caption{
Visual OOD and robustness generalization results over 10 trials.
All values are reported as percentages.
For cloth sweep, the score denotes full-task success.
For pick-and-place and cup stacking, the score denotes the average progress over ordered milestones.
All quantized methods use INT4 W4A16.
}
\label{tab:visual_ood_robustness}

\footnotesize
\setlength{\tabcolsep}{4pt}
\setlength{\aboverulesep}{1.5pt}
\setlength{\belowrulesep}{1.5pt}
\renewcommand{\arraystretch}{1.0}

\begin{tabularx}{\linewidth}{@{}lll*{5}{Y}@{}}
\toprule
Task & Shift & Method
& $\pi_0$
& $\pi_{0.5}$
& $\pi_0$-FAST
& SmolVLA
& GR00T N1.5 \\
\midrule

\multirow{9}{*}{Cloth sweep}
& \multirow{3}{*}{Unseen color}
& FP   & 100 & 100 & 80 & 80 & 80 \\
& & PTQ  & 100 & 100 & 70 & 70 & 60 \\
& & QAT  & 100   & 100 & 70& 80 & 80 \\
\cmidrule(lr){2-8}

& \multirow{3}{*}{Illumination}
& FP   & 100 & 100 & 70 & 100 & 70 \\
& & PTQ  & 100 & 50  & 50 & 90  & 60 \\
& & QAT  & 100  & 60  & 60 & 100  & 60 \\
\cmidrule(lr){2-8}

& \multirow{3}{*}{Sensor deg.}
& FP   & 80 & 100 & 60 & 50 & 60 \\
& & PTQ  & 50 & 30  & 30 & 40 & 30 \\
& & QAT  & 60 &  60 & 30 & 50 & 30 \\

\midrule

\multirow{12}{*}{Pick \& place}
& \multirow{3}{*}{Unseen background}
& FP   & 85 & 75 & 20 & 30 & 50 \\
& & PTQ  & 70 & 75 & 10 & 15 & 40 \\
& & QAT  & 75 & 75 & 15 & 25 & 50 \\
\cmidrule(lr){2-8}

& \multirow{3}{*}{Illumination}
& FP   & 70 & 70 & 20 & 60 & 60 \\
& & PTQ  & 60 & 55 & 10 & 35 & 55 \\
& & QAT  & 60 & 60 & 15 & 45 & 50 \\
\cmidrule(lr){2-8}

& \multirow{3}{*}{Sensor deg.}
& FP   & 55 & 80 & 10 & 65 & 40 \\
& & PTQ  & 45 & 10 & 0 & 0  & 25 \\
& & QAT  & 45 & 55 & 5 & 15 & 20 \\
\cmidrule(lr){2-8}

& \multirow{3}{*}{Distractor}
& FP   & 0  & 70 & 0 & 0 & 0 \\
& & PTQ  & 0  & 55 & 0 & 0 & 0 \\
& & QAT  & 0 & 60 & 0 & 0 & 0 \\

\midrule

\multirow{12}{*}{Cup stacking}
& \multirow{3}{*}{Unseen background}
& FP   & 70 & 85 & 20 & 5  & 25 \\
& & PTQ  & 35 & 45 & 0  & 5  & 25 \\
& & QAT  & 60 & 75 & 15 & 5 & 20 \\
\cmidrule(lr){2-8}

& \multirow{3}{*}{Illumination}
& FP   & 45 & 50 & 10 & 40 & 60 \\
& & PTQ  & 30 & 35 & 10 & 15 & 50 \\
& & QAT  & 35 & 40 & 10 & 25 & 55 \\
\cmidrule(lr){2-8}

& \multirow{3}{*}{Sensor deg.}
& FP   & 30 & 30 & 10 & 15 & 35 \\
& & PTQ  & 5  & 0  & 0 & 10 & 10 \\
& & QAT  & 20 & 10 & 10 & 10 & 10 \\
\cmidrule(lr){2-8}

& \multirow{3}{*}{Distractor}
& FP   & 10 & 80 & 0 & 0 & 0 \\
& & PTQ  & 0  & 30 & 0 & 0 & 0 \\
& & QAT  & 10 & 35 & 0 & 0 & 0 \\

\bottomrule
\end{tabularx}
\end{minipage}
\end{table*}

\begin{table}[!t]
\centering
\caption{Additional longer-trained SmolVLA results (\%).}
\label{app:tab:smolvla_longer_training}
\footnotesize
\setlength{\tabcolsep}{3pt}
\renewcommand{\arraystretch}{1.1}
\begin{tabular*}{\columnwidth}{@{\extracolsep{\fill}}llccc@{}}
\toprule
Task & Metric & FP & PTQ & QAT \\
\midrule
Sweep & Full & 100.0 & 86.7 & 90.0 \\
Pick/place & Full & 93.3 & 80.0 & 90.0 \\
Stack & Full & 46.7 & 26.7 & 36.7 \\
\midrule
Folding & Full & 0.0 & 0.0 & 0.0 \\
 & Progress & 30.0 & 5.0 & 20.0 \\
Clean desk & Full & 0.0 & 0.0 & 0.0 \\
 & Progress & 10.0 & 5.0 & 7.5 \\
\bottomrule
\end{tabular*}
\end{table}

\subsection{Full Semantic-Shift Results}
\label{app:semantic_summary}

Table~\ref{app:tab:semantic_results_full} reports the task-level results for object replacement, recipient replacement, added context, and null inputs. The first two pick-and-place conditions form the main paper's unseen-object aggregate. Added-context results average the pick-and-place and stacking conditions. All these task-level values use mean completion of the pick and place stages over 10 trials per model--method pair.

\paragraph{Primitive versus instruction-conditioned behavior.}
The main paper's role-reversal result measures three-stage primitive progress irrespective of which cup is assigned each role. The instruction-conditioned diagnostic additionally checked the specified role: only $\pi_0$ under FP and QAT received nonzero correct-role scores, while the other evaluated models and all PTQ variants received zero scores. This distinction is useful because successful-looking manipulation is not sufficient evidence of correct role grounding. The role-agnostic numerical results are reported in Table~\ref{tab:comparative_models}.

\paragraph{Added context and null inputs.}
The additional prompts are listed in Table~\ref{app:tab:shift_prompts}. Nonzero null-input progress reflects task-like action without a valid instruction, not instruction-conditioned task success. It is kept distinct from the main semantic averages. Together, the replacement, context, and role diagnostics distinguish manipulation capability from the stronger requirement of selecting behavior according to the language input.

\begin{table*}[t]
\centering
\caption{Full task-level semantic-shift and null-input diagnostic results (\%). Rows retain separate FP, PTQ, and QAT scores. Quantized methods use INT4 W4A16. Null-input values describe task-like progress, not success for a valid instruction.}
\label{app:tab:semantic_results_full}
\footnotesize
\setlength{\tabcolsep}{4pt}
\renewcommand{\arraystretch}{1.10}
\begin{tabular*}{\textwidth}{@{\extracolsep{\fill}}lllccccc@{}}
\toprule
Task & Condition & Method & $\pi_0$ & $\pi_{0.5}$ & $\pi_0$-FAST & SmolVLA & GR00T N1.5 \\
\midrule
Pick/place & Corn $\rightarrow$ carrot & FP & 90.0 & 90.0 & 10.0 & 45.0 & 60.0 \\
 &  & PTQ & 65.0 & 50.0 & 5.0 & 10.0 & 50.0 \\
 &  & QAT & 85.0 & 75.0 & 10.0 & 25.0 & 50.0 \\
\midrule
Pick/place & Plate $\rightarrow$ pot & FP & 55.0 & 85.0 & 5.0 & 35.0 & 45.0 \\
 &  & PTQ & 45.0 & 40.0 & 0.0 & 10.0 & 35.0 \\
 &  & QAT & 50.0 & 60.0 & 0.0 & 25.0 & 40.0 \\
\midrule
Pick/place & Added context & FP & 50.0 & 100.0 & 5.0 & 60.0 & 45.0 \\
 &  & PTQ & 50.0 & 50.0 & 0.0 & 30.0 & 50.0 \\
 &  & QAT & 40.0 & 75.0 & 10.0 & 50.0 & 45.0 \\
\midrule
Stack & Added context & FP & 40.0 & 40.0 & 5.0 & 0.0 & 40.0 \\
 &  & PTQ & 30.0 & 25.0 & 0.0 & 0.0 & 25.0 \\
 &  & QAT & 35.0 & 30.0 & 0.0 & 0.0 & 30.0 \\
\midrule
Pick/place & Null instruction & FP & 20.0 & 90.0 & 0.0 & 65.0 & 55.0 \\
 &  & PTQ & 0.0 & 30.0 & 0.0 & 0.0 & 35.0 \\
 &  & QAT & 10.0 & 45.0 & 0.0 & 30.0 & 40.0 \\
\midrule
Stack & Null instruction & FP & 30.0 & 45.0 & 0.0 & 0.0 & 20.0 \\
 &  & PTQ & 20.0 & 45.0 & 0.0 & 0.0 & 5.0 \\
 &  & QAT & 20.0 & 45.0 & 0.0 & 0.0 & 10.0 \\
\bottomrule
\end{tabular*}
\end{table*}


\subsection{Action-Side Reliability: Action Chunk Entropy}
\label{app:ace_details}

Task completion alone does not describe the concentration of a policy's
generated actions. We use Action Chunk Entropy (ACE)~\cite{romer2025fiper} as a complementary rollout-level
diagnostic. For an observation and language instruction at step $t$, let
$A_t^{(b)}=(a_{t,0}^{(b)},\ldots,a_{t,H-1}^{(b)})$ denote one sampled action
chunk of horizon $H$. We draw $B=10$ chunks using different sampling or
denoising seeds and form an empirical action distribution $\hat p_{t,h}$ at
each chunk offset $h$. The reported statistic sums entropy across chunk
offsets and averages over the evaluated rollout steps:
\begin{equation}
\overline{\mathrm{ACE}}
=
\frac{\displaystyle
\sum_{\tau\in\mathcal{E}}\sum_{t=1}^{T_\tau}
\sum_{h=0}^{H-1}\mathcal{H}(\hat p_{t,h})}
{\displaystyle\sum_{\tau\in\mathcal{E}}T_\tau},
\label{app:eq:ace}
\end{equation}
where $\mathcal{E}$ is the set of evaluation episodes and $T_\tau$ is the
number of evaluated steps in episode $\tau$. For continuous actions, the
entropy is computed from an estimated action distribution following the ACE
formulation. Lower values indicate more concentrated generated chunks under
the same evaluation protocol; the statistic is not normalized by $H$.

\begin{table*}[t]
\centering
\caption{Action-side diagnostic results for $\pi_{0.5}$.
Scores are percentages; ``Full'' denotes full task success and ``Milestone''
denotes milestone progress. ACE is reported separately from task score and
is not a success-rate estimate.}
\label{app:tab:ace_reliability}
\footnotesize
\setlength{\tabcolsep}{4pt}
\renewcommand{\arraystretch}{1.12}
\begin{tabular*}{\textwidth}{@{\extracolsep{\fill}}llcccccc@{}}
\toprule
\multirow{2}{*}{Task / condition} & \multirow{2}{*}{Score metric}
& \multicolumn{2}{c}{FP} & \multicolumn{2}{c}{PTQ} & \multicolumn{2}{c}{QAT} \\
\cmidrule(lr){3-4}\cmidrule(lr){5-6}\cmidrule(lr){7-8}
& & Score $\uparrow$ & ACE $\downarrow$ & Score $\uparrow$ & ACE $\downarrow$
& Score $\uparrow$ & ACE $\downarrow$ \\
\midrule
Cloth sweeping / ID & Full & 100.0 & 112.2 & 100.0 & 116.2 & 100.0 & 112.9 \\
Pick-and-place / ID & Full & 76.7 & 115.2 & 63.3 & 121.5 & 76.7 & 117.2 \\
Cup stacking / ID & Full & 83.3 & 123.9 & 46.7 & 131.5 & 83.3 & 125.2 \\
Pick-and-place / sensor degradation & Milestone & 80.0 & 189.6 & 10.0 & 204.5 & 55.0 & 198.3 \\
\bottomrule
\end{tabular*}
\end{table*}

Table~\ref{app:tab:ace_reliability} reports all four evaluated task--condition settings. PTQ increases ACE in each reported setting, while QAT
reduces it relative to PTQ. QAT nevertheless retains a higher ACE than FP,
including on ID pick-and-place and cup stacking, where the full-success
scores recover to the FP values. For sensor-degraded pick-and-place, QAT
improves milestone progress from 10.0 to 55.0, while ACE decreases from
204.5 to 198.3 and remains above FP's 189.6.

These observations illustrate why we do not treat ACE as a monotonic
predictor of success. A policy can complete a task despite a more dispersed
action distribution, and a concentrated distribution can still generate
incorrect actions. The residual ACE difference is an action-side diagnostic
that complements the task scores. Moreover, each policy is evaluated on its
own rollouts, so cross-policy ACE differences also reflect differences in
visited observations. They do not isolate a causal effect of quantization at
matched states or establish statistical significance.

\subsection{Visual-Side Reliability: Nuisance Mass Ratio}
\label{app:nmr_details}

We use Nuisance Mass Ratio (NMR) as a visual-side diagnostic of where
action-sensitive image evidence is located. For observation $o_t$ and
instruction $l$, let $S_t(r)$ denote perturbation-based action saliency in
region $r$, let $K_t^k$ be the set of the top-$k$ salient regions, and let
$M_t^{\mathrm{nuis}}$ denote the nuisance mask over task-irrelevant regions.
NMR measures the fraction of selected saliency mass inside that mask:
\begin{equation}
\mathrm{NMR}@k(o_t,l)
=
\frac{\sum_{r\in K_t^k\cap M_t^{\mathrm{nuis}}}S_t(r)}
{\sum_{r\in K_t^k}S_t(r)}.
\label{app:eq:nmr}
\end{equation}
Lower NMR indicates less measured action saliency on the designated nuisance
regions. NMR is used only for analysis, rather than as a training objective
or a substitute for task score.

\paragraph{Perturbation protocol.}
We use $k=10$. Policy inputs are resized to a $224\times224$ letterboxed
image and perturbed using a $16\times16$ Bernoulli mask grid. The mask ratio
is 0.30, with 100 interventions per frame. Perturbations use Gaussian blur
with kernel size 31 and $\sigma=8$. The saliency map measures how these
local image perturbations change the generated action. Dense evaluation
samples up to 50 gated frames per episode. The same prompt and
mask-generation pipeline is used across precision settings.

\paragraph{Task-relevant mask construction.}
In the real-world experiments, task-relevant masks are generated with
CLIPSeg~\cite{luddecke2022clipseg}, using
\texttt{CIDAS/clipseg-rd64-refined}. The default segmentation threshold is
0.30, with category-specific thresholds for action objects, support objects,
and robot regions. The pipeline also uses separate robot-region
thresholds for the base and wrist views. Table~\ref{app:tab:nmr_prompt_groups}
lists the twelve action/support prompt groups and their
view-specific thresholds. Action masks identify manipulated objects or
robot--object interaction regions; support masks identify receptacles, goal
regions, and task-relevant target areas.

\begin{table*}[t]
\centering
\caption{Task-specific action/support CLIPSeg prompt sets and thresholds used for NMR evaluation.}
\label{app:tab:nmr_prompt_groups}
\footnotesize
\setlength{\tabcolsep}{4pt}
\renewcommand{\arraystretch}{1.1}
\begin{tabularx}{\textwidth}{@{}llll>{\raggedright\arraybackslash}X@{}}
\toprule
Task & View & Mask type & Threshold & Prompt set \\
\midrule
Pick-and-place & Base & Action & 0.18
& target-object prompts: small yellow corn, yellow corn kernel, small corn on metal bowl, yellow corn kernels in bowl \\
Pick-and-place & Wrist & Action & 0.16
& target-object prompts: yellow corn cob only, yellow plastic corn kernel, small yellow corn on table, yellow corn cob tip only \\
Pick-and-place & Base & Support & 0.26
& receptacle prompts: metal bowl, metal plate, bowl rim, metal lip \\
Pick-and-place & Wrist & Support & 0.11
& receptacle prompts: metal bowl, metal plate, bowl rim, metal lip \\
\midrule
Cup stacking & Base & Action & 0.08
& manipulated-cup prompts: pink cup, red cup, pink top-view cup disk, red top-view cup disk, target cup \\
Cup stacking & Wrist & Action & 0.02
& manipulated-cup prompts: gripper-held cup, held pink cup, pink cup rim, red cup rim, cup in gripper \\
Cup stacking & Base & Support & 0.08
& goal-cup prompts: blue cup, cyan cup, blue cup opening, cyan cup opening, goal cup rim \\
Cup stacking & Wrist & Support & 0.02
& goal-cup prompts: blue cup rim, cyan cup rim, receiving cup opening, target stacking cup \\
\midrule
Clean desk & Base & Action & 0.22
& manipulated-item prompts: desk item, object on desk, trash item, small object on table, item to clean \\
Clean desk & Wrist & Action & 0.26
& manipulated-item prompts: object in gripper, held item, trash item, small desk object, item being picked \\
Clean desk & Base & Support & 0.24
& receptacle prompts: green basket on table, green plastic basket on desk, yellow box on desk, yellow trash box on table, open yellow bin on desk \\
Clean desk & Wrist & Support & 0.22
& receptacle prompts: green basket rim, green plastic bin opening, yellow box opening, yellow trash bin rim \\
\bottomrule
\end{tabularx}
\end{table*}

\begin{figure}[t]
\centering
\includegraphics[width=\columnwidth]{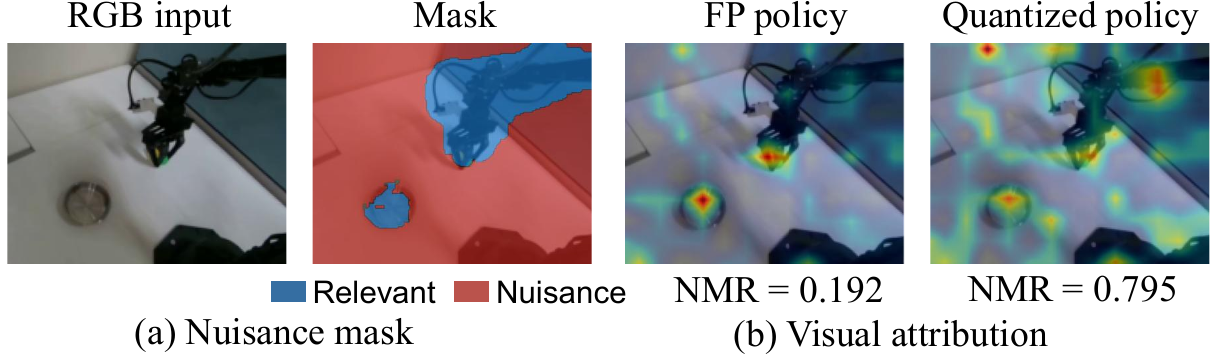}
\caption{Nuisance-mask and perturbation-attribution example used
to illustrate the NMR diagnostic. Numerical comparisons are reported in
Table~\ref{app:tab:nmr_reliability}.}
\label{app:fig:nmr_attribution}
\end{figure}

\paragraph{Results and interpretation.}
Figure~\ref{app:fig:nmr_attribution} illustrates the mask and attribution
construction. Across all four settings in
Table~\ref{app:tab:nmr_reliability}, PTQ has higher NMR than FP, and QAT
reduces NMR relative to PTQ while retaining a residual difference from FP.
For example, clean-desk NMR changes from 0.53 (FP) to 0.93 (PTQ) and 0.64
(QAT). We interpret this pattern as a relative attribution shift under the
specified masks and perturbations. It is not an absolute measure of visual
grounding quality, and it does not by itself identify the cause of a task
failure. Segmentation, perturbation design, selected frames, and the
observations visited by each policy affect the diagnostic. In particular,
the comparisons on policy-specific rollouts do not hold observations fixed.

\subsection{Representative Trajectory Comparison}
\label{app:trajectory_comparison}

Figure~\ref{app:fig:trajectory_comparison} complements the ACE and NMR
statistics with representative cup-stacking trajectories. The expert and FP
trajectories remain close in this comparison, whereas quantized trajectories
deviate near the placement stage. This qualitative comparison cannot
establish how often such drift occurs or distinguish action dispersion from
attribution changes as its cause. Together with the quantitative task results, these
diagnostics motivate examining outcomes over complete rollouts: action-level
agreement, action concentration, visual attribution, and successful task
completion characterize different aspects of policy behavior.

\subsection{Model-Level Robustness Profile}
\label{app:model_level_profile}

Table~\ref{app:tab:model_level_profile} summarizes each model across three evaluation axes using the task-level results reported in this paper; it introduces no additional experiments. ID averages full success over the three short tasks. Long horizon averages the progress scores for folding and desk cleaning, preserving partial-completion information; this differs from the long-horizon full-success average in Fig.~\ref{fig:difficulty_visualgen}(a). Visual OOD gives equal weight to each of the 11 evaluated task--condition pairs: three each for appearance, illumination, and sensor degradation, and two for distractors. It therefore differs from an equally weighted average over the four condition-level means.

\clearpage
\twocolumn[
\begin{minipage}{\textwidth}
\settablecaption
\begin{minipage}[t]{0.48\textwidth}
\vspace{0pt}
\centering
\caption{NMR comparisons for $\pi_{0.5}$. Lower values
indicate less saliency mass in designated nuisance regions. }
\label{app:tab:nmr_reliability}
\footnotesize
\setlength{\tabcolsep}{4pt}
\renewcommand{\arraystretch}{1.1}
\begin{tabularx}{\columnwidth}{@{}>{\raggedright\arraybackslash}Xccc@{}}
\toprule
Task / condition & FP & PTQ & QAT \\
\midrule
Pick-and-place / ID & 0.36 & 0.85 & 0.42 \\
Pick-and-place / OOD & 0.51 & 0.91 & 0.60 \\
Cup stacking / ID & 0.45 & 0.83 & 0.53 \\
Clean desk / Long-horizon & 0.53 & 0.93 & 0.64 \\
\bottomrule
\end{tabularx}

\par\vspace{9pt}
\begingroup
\setfigurecaption
\centering
\includegraphics[width=0.75\linewidth]{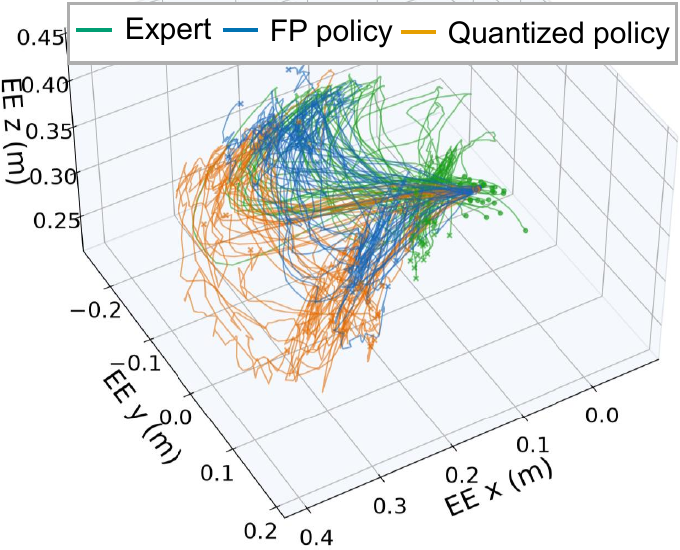}
\caption{Representative cup-stacking trajectory comparison. The quantized rollout deviates near placement;
this example is qualitative rather than an aggregate performance metric.}
\label{app:fig:trajectory_comparison}

\endgroup
\end{minipage}\hfill
\begin{minipage}[t]{0.48\textwidth}
\vspace{0pt}
\centering
\caption{Derived model-level robustness profiles (\%). Parentheses show the drop from FP in percentage points, computed before rounding.}
\label{app:tab:model_level_profile}
\footnotesize
\setlength{\tabcolsep}{2pt}
\renewcommand{\arraystretch}{1.12}
\begin{tabular*}{\columnwidth}{@{\extracolsep{\fill}}llccc@{}}
\toprule
Model & Axis & FP & PTQ & QAT \\
\midrule
\multirow{3}{*}{$\pi_0$}
 & ID & 81.1 & 71.1 (10.0) & 83.3 ($-2.2$) \\
 & Long & 73.8 & 31.3 (42.5) & 71.3 (2.5) \\
 & Visual & 58.6 & 45.0 (13.6) & 51.4 (7.3) \\
\midrule
\multirow{3}{*}{$\pi_{0.5}$}
 & ID & 86.7 & 70.0 (16.7) & 86.7 (0.0) \\
 & Long & 65.0 & 27.5 (37.5) & 46.3 (18.8) \\
 & Visual & 76.4 & 44.1 (32.3) & 57.3 (19.1) \\
\midrule
\multirow{3}{*}{$\pi_0$-FAST}
 & ID & 50.0 & 38.9 (11.1) & 47.8 (2.2) \\
 & Long & 13.8 & 1.3 (12.5) & 5.0 (8.8) \\
 & Visual & 27.3 & 16.4 (10.9) & 20.9 (6.4) \\
\midrule
\multirow{3}{*}{SmolVLA}
 & ID & 48.9 & 30.0 (18.9) & 40.0 (8.9) \\
 & Long & 7.5 & 0.0 (7.5) & 0.0 (7.5) \\
 & Visual & 40.5 & 25.5 (15.0) & 32.3 (8.2) \\
\midrule
\multirow{3}{*}{GR00T N1.5}
 & ID & 56.7 & 44.4 (12.2) & 50.0 (6.7) \\
 & Long & 40.0 & 26.3 (13.8) & 30.0 (10.0) \\
 & Visual & 43.6 & 32.3 (11.4) & 34.1 (9.5) \\
\bottomrule
\end{tabular*}
\par\smallskip
\begin{minipage}{\columnwidth}
\footnotesize
ID: short-task full success; Long: long-horizon progress; Visual: visual-OOD task score. Quantized policies use INT4 W4A16. A negative drop means that the method exceeds its FP reference on that aggregate.
\end{minipage}

\end{minipage}
\end{minipage}
\vspace{12pt}]
\section{FP Rollout Augmentation Bound}
\label{app:fp_rollout_bound}
\label{app:finite_horizon_proof}

This appendix develops the advantage--mismatch argument behind FP rollout augmentation and proves Eq.~\eqref{eq:deployment_policy_bound}. The analysis follows the performance-difference identity and a conservative state-distribution bound~\cite{kakade2002approximately,schulman2015trpo}; it is not a new policy-improvement theorem or an optimality guarantee for QCRL. An auxiliary source is useful when it supplies beneficial actions at states relevant to the quantized policy, while remaining sufficiently close for those actions to be extracted. We use a finite horizon with $\gamma=1$, consistently with the main analysis and advantage-labeling procedure.

\subsection{Setup and Assumptions}
\label{app:bound_setup}

Let $\pi_Q^0$ be the initial PTQ policy and $\pi_s$ an auxiliary source, such as its FP counterpart or an external policy. All policies act in the same horizon-$T$ Markov process, with a common initial distribution, transition dynamics, and reward function. The state includes the instruction, sufficient history, and milestone-completion status. Termination, including a time limit, is followed by an absorbing state with zero subsequent rewards; the actual terminal reward is retained. This permits trajectories of different lengths to be compared over a common horizon without changing their returns.

Write $d_t^s$ and $d_t^0$ for the source and PTQ state distributions at time $t$, respectively, and let $J_T$ be expected undiscounted return. The exact time-dependent PTQ advantage is
\begin{equation}
\label{eq:app_exact_advantage}
A_t^0(s,a)=Q_t^{\pi_Q^0}(s,a)-V_t^{\pi_Q^0}(s).
\end{equation}
Assume $|A_t^0(s,a)|\leq A_{\max}$ for every $t,s,a$. Define the uniform source distance and its expected advantage at a state as
\begin{equation}
\label{eq:app_source_definitions}
\begin{aligned}
\alpha_s&=\sup_{t,s}\mathrm{TV}\!\left(
\pi_{s,t}(\cdot\mid s),\pi_{Q,t}^0(\cdot\mid s)\right),\\
g_{s,t}(s)&=\mathbb E_{a\sim\pi_{s,t}(\cdot\mid s)}[A_t^0(s,a)].
\end{aligned}
\end{equation}
Here, TV is half the $L_1$ distance between probability distributions. For stochastic action heads, the distribution includes action-token sampling or diffusion/flow sampling noise. For deterministic continuous-action policies, distinct point masses can have TV equal to one even when their actions are numerically close. Thus, a small empirical action distance does not establish a small uniform TV distance. QCRL does not optimize or estimate $\alpha_s$ directly.

\subsection{Finite-Horizon Surrogate and Occupancy Bound}
\label{app:finite_horizon_surrogate}

The performance-difference identity gives
\begin{equation}
\label{eq:app_finite_performance_difference}
J_T(\pi_s)-J_T(\pi_Q^0)
=\sum_{t=1}^{T}\mathbb E_{s\sim d_t^s}[g_{s,t}(s)].
\end{equation}
This evaluates source actions under the states visited by the source. To express their benefit at PTQ states, define the surrogate
\begin{equation}
\label{eq:app_source_surrogate}
S_s=\sum_{t=1}^{T}\mathbb E_{s\sim d_t^0}[g_{s,t}(s)].
\end{equation}
Adding and subtracting this quantity separates action benefit from the change in state distribution:
\begin{equation}
\label{eq:app_occupancy_residual}
\begin{aligned}
J_T(\pi_s)-J_T(\pi_Q^0)&=S_s+\Delta_s,\\
\Delta_s&=\sum_{t=1}^{T}
\left(\mathbb E_{d_t^s}g_{s,t}-\mathbb E_{d_t^0}g_{s,t}\right).
\end{aligned}
\end{equation}

The shared initial distribution gives $d_1^s=d_1^0$. A maximal coupling of the two action distributions disagrees with probability at most $\alpha_s$ whenever the policies are in the same state. Coupling the common dynamics until this first disagreement and taking a union bound gives
\begin{equation}
\label{eq:app_state_occupancy_bound}
\|d_t^s-d_t^0\|_1\leq2(t-1)\alpha_s.
\end{equation}
The bound remains valid when its right-hand side exceeds two, although it then becomes loose.

For a bounded function $f$, the TV expectation inequality is $|\mathbb E_p f-\mathbb E_q f|\leq2\|f\|_\infty\mathrm{TV}(p,q)$. Since the exact advantage satisfies $\mathbb E_{a\sim\pi_{Q,t}^0}A_t^0(s,a)=0$, applying this inequality at each state yields
\begin{equation}
\label{eq:app_expected_advantage_bound}
|g_{s,t}(s)|\leq2A_{\max}\alpha_s.
\end{equation}
Combining Eqs.~\eqref{eq:app_state_occupancy_bound} and~\eqref{eq:app_expected_advantage_bound} therefore gives
\begin{equation}
\begin{aligned}
|\Delta_s|
&\leq\sum_{t=1}^{T}\|g_{s,t}\|_\infty
\|d_t^s-d_t^0\|_1\\
&\leq4A_{\max}\alpha_s^2\sum_{t=1}^{T}(t-1)\\
&=2A_{\max}T(T-1)\alpha_s^2.
\end{aligned}
\end{equation}
The resulting source-policy bound is
\begin{equation}
\label{eq:app_fp_source_bound}
\begin{aligned}
J_T(\pi_s)-J_T(\pi_Q^0)
\geq S_s-2A_{\max}T(T-1)\alpha_s^2.
\end{aligned}
\end{equation}
Its quadratic mismatch term follows from bounding both the state-distribution change and the expected advantage using the same policy distance.

\subsection{Extension to Reference-Conditioned Policy Extraction}
\label{app:reference_extraction_proof}

QCRL does not directly deploy the rollout source $\pi_s$. It trains a conditional quantized policy on demonstrations, FP rollouts, and quantized rollouts, then fixes $I_t=\mathrm{pos}$ and $q_t=\mathtt{ref}$ at deployment. The reference group includes demonstrations as well as FP experience. Accordingly, the idealized positive-reference policy $\pi_r^+$ in Section~\ref{sec:source_selection_analysis} need not equal the raw FP rollout policy.

Assume $\pi_r^+$ is defined at every state. Use the uniform approximation error $\epsilon_{\mathrm{BC}}$, reference--PTQ distance $\alpha_r$, and PTQ-state reference benefit $S_r$ defined in Eq.~\eqref{eq:source_advantage_tv}. Writing $\pi_d=\pi_{\mathrm{deploy}}$, the triangle inequality gives
\begin{equation}
\label{eq:app_deployment_tv_bound}
\begin{aligned}
\delta_d&:=\sup_{t,s}\mathrm{TV}\!\left(
\pi_{d,t}(\cdot\mid s),\pi_{Q,t}^0(\cdot\mid s)\right)\\
&\leq\alpha_r+\epsilon_{\mathrm{BC}}.
\end{aligned}
\end{equation}
The same TV expectation inequality bounds the effect of imperfect reference extraction at each state:
\begin{equation}
\label{eq:app_reference_advantage_approximation}
\begin{aligned}
\mathbb E_{a\sim\pi_{d,t}}[A_t^0(s,a)]
&\geq\mathbb E_{a\sim\pi_{r,t}^+}[A_t^0(s,a)]\\
&\quad-2A_{\max}\epsilon_{\mathrm{BC}}.
\end{aligned}
\end{equation}
Both action expectations condition on the same state $s$. Let $S_d$ denote the surrogate in Eq.~\eqref{eq:app_source_surrogate} for $\pi_d$. Summing over $t$ under $d_t^0$ gives
\begin{equation}
\label{eq:app_deployment_surrogate_bound}
S_d\geq S_r-2TA_{\max}\epsilon_{\mathrm{BC}}.
\end{equation}
Applying Eq.~\eqref{eq:app_fp_source_bound} with $\pi_s=\pi_d$, then substituting Eqs.~\eqref{eq:app_deployment_tv_bound} and~\eqref{eq:app_deployment_surrogate_bound}, proves Eq.~\eqref{eq:deployment_policy_bound}. The learned policy therefore inherits the source advantage--mismatch trade-off, with an additional approximation term and an increased state-distribution penalty.

\subsection{Interpretation and Scope}
\label{app:source_bound_interpretation}

For the auxiliary source itself, the lower bound is positive when $S_s>2A_{\max}T(T-1)\alpha_s^2$: beneficial actions at PTQ states must outweigh the possible effect of changing the state distribution. High return under the source's own occupancy does not by itself establish this condition. For the learned policy, the reference benefit must additionally exceed the extraction error in Eq.~\eqref{eq:deployment_policy_bound}. The horizon dependence indicates why small deviations can become consequential in longer executions.

The corresponding FP policy is a practical source candidate because it shares the backbone, action representation, and task training of the PTQ policy while providing better rollout outcomes in our evaluated settings. An external expert can have high task performance yet produce substantially different actions. These observations motivate source selection, but do not establish the theorem's uniform assumptions. Similarly, source conditioning makes selected reference behavior separately representable; the bound does not prove that conditioning minimizes approximation error or is optimal.

The analysis uses exact PTQ advantages and a reference policy defined throughout the state space. Training instead uses fitted advantages and a finite offline dataset. Neither these fitted labels nor the action-distance diagnostics below certify $S_r$, $\alpha_r$, or $\epsilon_{\mathrm{BC}}$. The bound is a design guide, while Section~\ref{sec:qcrl_ablation_results} evaluates the contributions of FP rollout augmentation, progress-based rewards, and source conditioning experimentally.

\subsection{Empirical Source-Distance Diagnostics}
\label{app:source_distance_details}

Table~\ref{tab:appendix_source_distance_full} gives the complete action-space comparison underlying Table~\ref{tab:source_diagnostics}(c). Here, $\pi_Q^0$ is the initial quantized $\pi_{0.5}$ policy, $\pi_{\mathrm{FP}}^0$ is its corresponding FP counterpart, and $\pi_{\mathrm{ext}}$ is a separately trained external $\pi_0$ policy for the same robot setting. Each distance is the $L_2$ norm between normalized action chunks, evaluated on the same 100 observations from $\mathcal D_Q$.

\begin{table}[t]
\centering
\caption{Complete rollout-source action-distance comparison. Entries are mean $\pm$ observation SD over 100 shared PTQ observations. Within-policy rows use independent seeds and characterize sampling variability; they are not guaranteed lower bounds on between-policy distances.}
\label{tab:appendix_source_distance_full}
\footnotesize
\setlength{\tabcolsep}{3pt}
\renewcommand{\arraystretch}{1.10}
\begin{tabular*}{\columnwidth}{@{\extracolsep{\fill}}lcc@{}}
\toprule
Pair & Matched-noise $L_2$ & Fresh-noise $L_2$\\
\midrule
$\pi_{\mathrm{FP}}^0\leftrightarrow\pi_Q^0$
& $0.87\pm0.21$ & $1.38\pm0.46$\\
$\pi_{\mathrm{FP}}^0\leftrightarrow\pi_{\mathrm{FP}}^0$
& --- & $1.06\pm0.39$\\
$\pi_Q^0\leftrightarrow\pi_Q^0$
& --- & $1.14\pm0.40$\\
\midrule
$\pi_{\mathrm{ext}}\leftrightarrow\pi_Q^0$
& $18.07\pm3.57$ & $18.37\pm3.82$\\
$\pi_{\mathrm{ext}}\leftrightarrow\pi_{\mathrm{FP}}^0$
& $18.11\pm3.54$ & $18.37\pm3.78$\\
$\pi_{\mathrm{ext}}\leftrightarrow\pi_{\mathrm{ext}}$
& --- & $11.41\pm4.11$\\
\bottomrule
\end{tabular*}
\end{table}

Matched noise uses the same Gaussian seed for both policies, comparing their outputs under a shared noise sample. Fresh noise uses independent seeds, so the distance also includes sampling stochasticity. Within-policy fresh-noise comparisons characterize the variability introduced by this sampling procedure.

The FP counterpart's matched-noise distance to PTQ is $0.87\pm0.21$; its fresh-noise distance is on the scale of the FP and PTQ within-policy variability. The external policy's matched-noise distance to PTQ is $18.07\pm3.57$, approximately $20.8\times$ larger. These observations support the corresponding FP policy as a nearby source in the measured action space. They do not estimate worst-case TV distance, establish state-space coverage, or directly measure the distance between the learned positive-reference policy and PTQ. The diagnostics are not used as an optimization objective.

\end{document}